%% file: ms.tex
\documentclass[10pt,twocolumn,letterpaper]{article}

\usepackage[pagenumbers]{cvpr} 

\usepackage{graphicx}
\usepackage{amsmath}
\usepackage{amssymb}
\usepackage{booktabs}

\usepackage[utf8]{inputenc} 
\usepackage[T1]{fontenc}    
\usepackage{url}            
\usepackage{booktabs}       
\usepackage{amsfonts}       
\usepackage{nicefrac}       
\usepackage{microtype}

\usepackage{mathtools}
\usepackage{caption}
\usepackage{subcaption}
\usepackage[table]{xcolor}
\usepackage{multirow}
\usepackage{overpic}
\usepackage{fdsymbol}
\usepackage{arydshln}

\usepackage{lipsum}
\newcommand\blfootnote[1]{%
  \begingroup
  \renewcommand\thefootnote{}\footnote{#1}%
  \addtocounter{footnote}{-1}%
  \endgroup
}

\definecolor{UFM}{HTML}{FC7319}
\definecolor{SURFM}{HTML}{7E0F7E}
\definecolor{WFM}{HTML}{4DBEEE}
\definecolor{DiffNet}{HTML}{010101}
\definecolor{DS}{HTML}{EDCA20}
\definecolor{NM}{HTML}{0072BD}
\definecolor{CZO}{HTML}{77DF30}
\definecolor{UDM}{HTML}{016C01}
\definecolor{SyNoRiM}{HTML}{D62FF5}
\definecolor{Ours}{HTML}{FE2801}

\usepackage{pifont}
\newcommand{\cmark}{\ding{51}}
\newcommand{\xmark}{\ding{55}}

\usepackage[capitalize]{cleveref}
\crefname{section}{Sec.}{Secs.}
\Crefname{section}{Section}{Sections}
\Crefname{table}{Table}{Tables}
\crefname{table}{Tab.}{Tabs.}

\def\cvprPaperID{2031} 
\def\confName{CVPR}
\def\confYear{2023}

\begin{document}

\include{commands}

\title{G-MSM: Unsupervised Multi-Shape Matching with Graph-based Affinity Priors}

\author{Marvin Eisenberger, Aysim Toker, Laura Leal-Taixé$^{\dagger}$, Daniel Cremers\\[5pt]
Technical University of Munich
}
\maketitle

\blfootnote{$^{\dagger}$ Currently at NVIDIA}

\begin{abstract}
We present G-MSM (\textbf{G}raph-based \textbf{M}ulti-\textbf{S}hape \textbf{M}atching), a novel unsupervised learning approach for non-rigid shape correspondence. Rather than treating a collection of input poses as an unordered set of samples, we explicitly model the underlying shape data manifold. To this end, we propose an adaptive multi-shape matching architecture that constructs an affinity graph on a given set of training shapes in a self-supervised manner. The key idea is to combine putative, pairwise correspondences by propagating maps along shortest paths in the underlying shape graph. During training, we enforce cycle-consistency between such optimal paths and the pairwise matches which enables our model to learn topology-aware shape priors. We explore different classes of shape graphs and recover specific settings, like template-based matching (star graph) or learnable ranking/sorting (TSP graph), as special cases in our framework. Finally, we demonstrate state-of-the-art performance on several recent shape correspondence benchmarks, including real-world 3D scan meshes with topological noise and challenging inter-class pairs.
\end{abstract}

\begin{figure}
\begin{subfigure}[b]{0.61\linewidth}
    \centering
    \includegraphics[width=\linewidth]{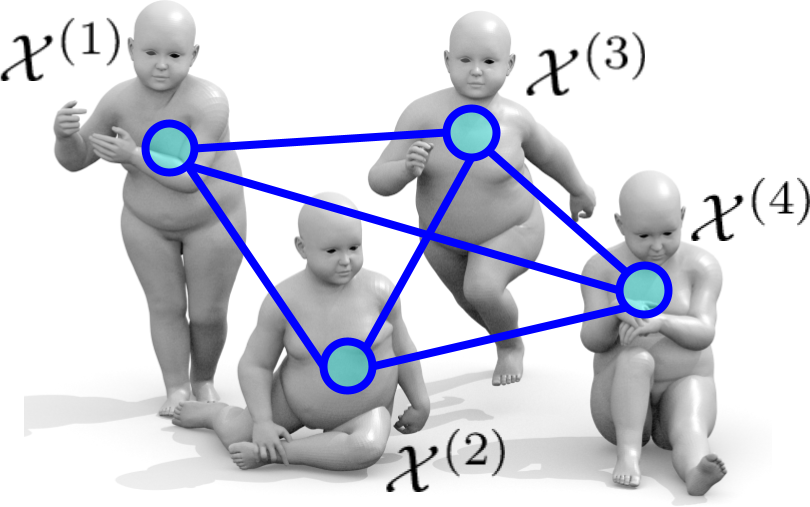}
    \vspace{-10pt}
    \label{fig:teas_a}
    \caption{Shape graph $\cG$}
\end{subfigure}
\begin{subfigure}[b]{0.37\linewidth}
    \centering
    \includegraphics[width=\linewidth]{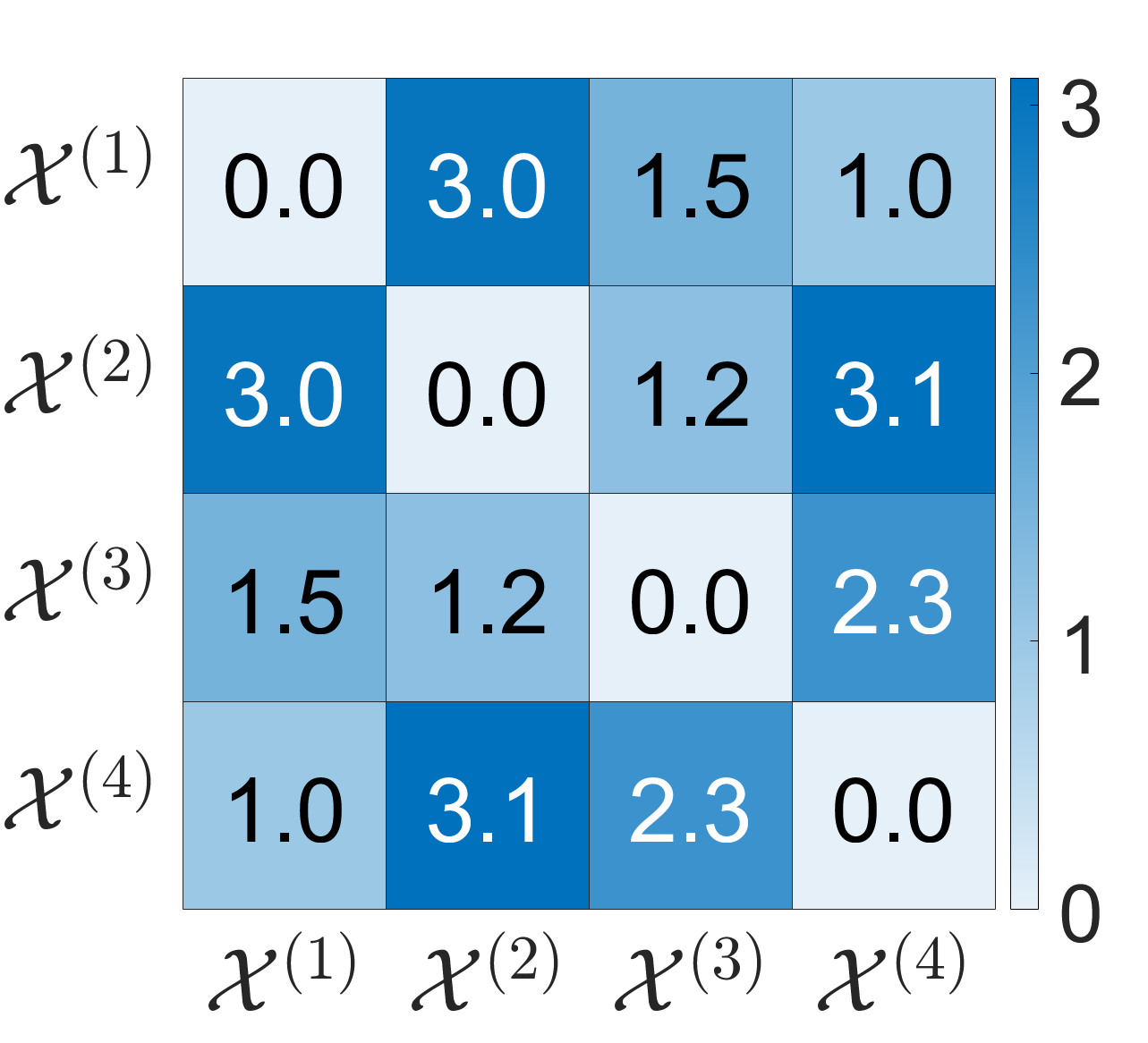}
    \vspace{-10pt}
    \label{fig:teas_b}
    \caption{Affinity edge weights}
\end{subfigure}
\begin{subfigure}[b]{\linewidth}
    \centering
    \vspace{10pt}
    \begin{overpic}
    [width=\linewidth]{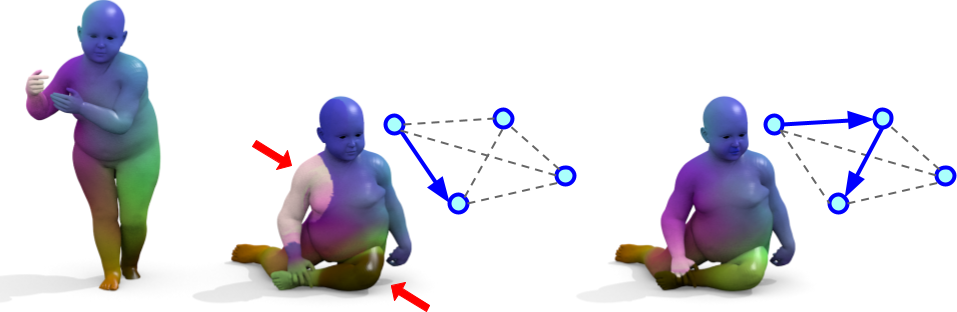}
    \put(15,28){Source}
    \put(41,26){$\mPi^{(1,2)}$}
    \put(74,26){$\mPi^{(1,3)}\circ\mPi^{(3,2)}$}
    \end{overpic}
    \vspace{-10pt}
    \label{fig:teas_c}
    \caption{Putative correspondences $\mPi^{(1,2)}$ vs multi-matching $\mPi^{(1,3)}\circ\mPi^{(3,2)}$}
\end{subfigure}
\caption{For a given collection of 3D meshes $\{\cXi|1\leq i\leq N\}$, (i) our method constructs, in a fully unsupervised manner, a shape graph $\cG$ which approximates the underlying shape data manifold. (ii) Its edge weights (affinity scores) are derived from a putative pairwise correspondence loss signal. (iii) During training, we enforce cycle-consistency by propagating maps along shortest paths in the graph $\cG$. As shown for the sample pair above $(\cX^{(1)}$, $\cX^{(2)})$, the resulting multi-matching $\mPi^{(1,3)}\circ\mPi^{(3,2)}$ is significantly more accurate than the pairwise map $\mPi^{(1,2)}$. 
}
\label{fig:teas}
\end{figure}

\section{Introduction}\label{sec:introduction}

Shape matching of non-rigid object categories is a central problem in 3D computer vision and graphics that has been studied extensively over the last few years.
Especially in recent times, there is a growing demand for such algorithms as 3D reconstruction techniques and affordable scanning devices become increasingly powerful and broadly available. Classical shape correspondence approaches devise axiomatic algorithms that make specific assumptions about the resulting maps, such as near-isometry, area preservation, approximate rigidity, bounded distortion, or commutativity with the intrinsic Laplacian. In contrast, real-world scan meshes are often subject to various types of noise, including topological changes~\cite{dyke2019shrec,laehner2016shrec}, partial views~\cite{attaiki2021dpfm}, general non-isometric deformations~\cite{dyke2020shrec,zuffi20173d}, objects in clutter~\cite{cosmo2016matching}, and varying data representations~\cite{sharp2022diffusionnet}. 
In this work, we address several of the aforementioned challenges and demonstrate that our proposed method achieves improved stability for a number of 3D scan mesh datasets. 

The majority of existing deep learning methods for shape matching~\cite{attaiki2021dpfm,donati2020deep,eisenberger2020deep,eisenberger2021neuromorph,halimi2019unsupervised,marin2020correspondence,roufosse2019unsupervised,sharma2020weakly} treat a given set of meshes as an unstructured collection of poses. During training, random pairs of shapes are sampled for which a neural network is queried and a pairwise matching loss is minimized. While this approach is straightforward, it often fails to recognize commonalities and context-dependent patterns which only emerge from analyzing the shape collection as a whole. \emph{Not all samples of a shape collection are created equal.} In most cases, some pairs of poses are much closer than others. Maps between similar geometries are inherently correlated and convey relevant clues to one another. This is particularly relevant for challenging real-world scenarios, where such redundancies can help disambiguate noisy geometries, non-isometric deformations, and topological changes. 
The most common approach of existing multi-matching methods is to learn a canonical embedding per pose, either in the spatial~\cite{cao2022unsupervised} or Laplace-Beltrami frequency domain~\cite{huang2020consistent,huang2022multiway}. This incentivizes the resulting matches to be consistent under concatenation. However, such approaches are in practice still trained in a fully pairwise manner for ease of training. Furthermore, relying on canonical embeddings can lead to limited generalization for unseen test poses. Concrete approaches often assume a specific mesh resolution and nearly-isometric poses~\cite{cao2022unsupervised}, or require an additional fine-tuning optimization at test time~\cite[Sec. 5]{huang2022multiway}.

Rather than interpreting a given training set as a random, unstructured collection of shapes, our approach explicitly models the underlying shape manifold.
To this end, we define an affinity graph $\cG$ on the set of input shapes whose edge weights (i.e. affinity scores) are informed by the outputs of a pairwise matching module. We then devise a novel adaptive multi-matching architecture that propagates matches along shortest paths in the underlying shape graph $\cG$.
The resulting maps are topology-aware, i.e., informed by geometries from the whole shape collection. An example is shown in~\Cref{fig:teas}, where the multi-matching $\mPi^{(1,3)}\circ\mPi^{(3,2)}$ obtained by our approach is significantly more accurate than the naive, pairwise map $\mPi^{(1,2)}$. 
During training, we promote cycle-consistency of shortest paths in the shape graph. In summary, our contributions are as follows:
\begin{enumerate}
    \item Introduce the notion of an edge-weighted, undirected shape graph $\cG$ to approximate the underlying data manifold for an unordered collection of 3D meshes.
    \item Propose a novel, adaptive multi-shape matching approach that enforces cycle-consistency for optimal paths in the shape graph $\cG$ in a self-supervised manner.
    \item Demonstrate state-of-the-art performance for a range of challenging non-rigid matching tasks, including non-isometric matching due to topological noise~\cite{dyke2019shrec,laehner2016shrec} and inter-class pairs~\cite{dyke2020shrec,zuffi20173d}.
\end{enumerate}

\section{Related work}\label{sec:relatedwork}

\paragraph{Axiomatic correspondence methods}

Shape matching is an extensively studied topic with a variety of different approaches and methodologies. We summarize references relevant to our approach here and refer to recent surveys~\cite{sahilliouglu2019recent,vankaick11correspsurvey} for a more complete picture.
Classical methods for non-rigid matching often devise optimization-based approaches that minimize some type of distortion metric~\cite{bronstein2006generalized,kernel17,rodola2014dense,windheuser2011large}. A common prerequisite of many such methods is the extraction of hand-crafted local descriptors that are, in approximation, preserved under non-rigid shape deformations. Common definitions include histogram-based statistics~\cite{tombari2010SHOT} or fully intrinsic features based on the eigenfunctions of the Laplace-Beltrami operator~\cite{aubry2011WKS,rustamov2007laplace,sun2009concise}. 
Over the last few years, functional maps \cite{ovsjanikov2012functional} have become a central paradigm in shape matching. The core idea is to reframe the pairwise matching task from functions (points to points) to functionals (functions to functions). There are several extensions of the original framework to allow for partial matching~\cite{litany2017fullyspectral,litany2016puzzles,rodola2016partial}, orientation preservation~\cite{ren2018orientation}, iterative map upsampling~\cite{eisenberger2020smooth,melzi2019zoomout} and conformal maps~\cite{donati2021complex}. Our approach utilizes functional maps as a fundamental building block within the differentiable matching layer.

\paragraph{Learning-based methods}

More recently, several approaches emerged that aim at extending the power of deep feature learning to deformable 3D shapes. Many such methods fall under the umbrella term `geometric deep learning'~\cite{bronstein2017geometric}, with analogous applications on different classes of non-Euclidean data like graphs or general manifold data.
One class of approaches are charting-based methods~\cite{boscaini2016learning,masci2015geodesic,monti2017geometric,poulenard2018multi,Wiersma2020} which imitate convolutions in Euclidean space with parameterized, intrinsic patch operators. Likewise,~\cite{sharp2022diffusionnet} proposed a learnable feature refinement module based on intrinsic heat diffusion. We employ the latter as the backbone of our pipeline.

The pioneering work of \cite{litany2017deep} proposes a differentiable matching layer based on functional maps~\cite{ovsjanikov2012functional}, in combination with a deep feature extractor with several consecutive ResNet layers~\cite{he2016deep}. Numerous extensions of this paradigm were proposed over the last few years to allow for unsupervised loss functions \cite{halimi2019unsupervised,roufosse2019unsupervised}, learnable basis functions~\cite{marin2020correspondence}, point cloud feature extractors~\cite{donati2020deep,huang2022multiway,sharma2020weakly} or partial data~~\cite{attaiki2021dpfm}.
Similarly, DeepShells~\cite{eisenberger2020deep} learns functional maps in an end-to-end trainable, hierarchical multi-scale pipeline. We adapt parts of its differentiable matching layer in our network.
Other approaches~\cite{bhatnagar2020loopreg,groueix20183dcoded,marin2018farm} learn correspondences for a specific class of deformable objects by including additional domain knowledge, like a deformable human model~\cite{SMPL:2015}. Finally,~\cite{eisenberger2021neuromorph} jointly learns to predict correspondences and a smooth interpolation between pairs of shapes.

\paragraph{Multi-shape matching}
Classical axiomatic multi-shape matching approaches devise optimization-based pipelines that enforce cycle-consistent maps. Specific solutions include semidefinite programming~\cite{huang2013consistent}, convex relaxations of the corresponding quadratic assignment problem \cite{kezurer2015tight}, graph cuts~\cite{schmidt2007intrinsic}, as well as evolutionary game theory~\cite{cosmo2017consistent}. Such optimization-based approaches are computationally costly and therefore limited to matching sparse landmarks.
Furthermore, there are a number of optimization frameworks that compute synchronized, cycle-consistent functional maps~\cite{gao2021isometric,huang2020consistent,huq2021riemannian}. Notably, such approaches are often limited to nearly-isometric poses~\cite{gao2021isometric,huq2021riemannian} or require high-quality initializations~\cite{huang2020consistent}. 
More recent learning-based approaches promote cycle-consistency by predicting a canonical embedding for each observed pose~\cite{cao2022unsupervised,huang2022multiway}. However, obtaining stable embeddings is often difficult when generalizing to unseen test poses. Moreover, such approaches assume a specific mesh resolution and nearly-isometric poses~\cite{cao2022unsupervised} or require an additional fine-tuning optimization at test time to obtain canonical embeddings~\cite[Sec. 5]{huang2022multiway}.

\section{Method}\label{sec:method}

\subsection{Problem formulation}\label{sec:problemformulation}

In the following, we consider a collection of 3D shapes $\cS=\bigl\{\cX^{(1)},\dots,\cX^{(N)}\bigr\}$ from non-rigidly deformable shape categories.
Each such shape $\cXi$ is a discretized approximation of a 2D Riemannian manifold, embedded in $\bbR^3$. Specifically, we define $\cXi=\bigl(\mVi,\mTi\bigr)$, where $\mVi\in\bbR^{m\times 3}$ and $\mTi\subset \mVi\times\mVi\times\mVi$ are sets of vertices and triangular faces, respectively. The goal is then to construct an algorithm that computes dense correspondence mappings $\mPiij$ between any two surfaces $\cXi$ and $\cXj$ from the shape collection $\cS$. Specifically, such correspondences are represented by sparse assignment matrices $\mPiij\in\{0,1\}^{m\times n}$, where ${\mPiij}\mathbf{1}_n=\mathbf{1}_m$ and $\mPiij_{i',j'}=1$ indicates a match between the $i'$-th vertex of $\cXi$ and the $j'$-th vertex of $\cXj$. 

\paragraph{Scope} Our method is unsupervised and thereby requires no additional inputs, like landmark annotations or ground-truth correspondences, beyond the raw input geometries $\cXi$. Following similar approaches in this line of work~\cite{bhatnagar2020loopreg,eisenberger2021neuromorph,marin2020correspondence,sharma2020weakly}, we assume that the shapes $\cX^{(1)},\dots,\cX^{(N)}$ have an approximately canonical orientation. In the literature, this setting is commonly referred to as `weakly supervised', see~\cite{sharma2020weakly} and later~\cite{eisenberger2021neuromorph}. 
Existing approaches often make additional assumptions about the input data $\cS$, focusing on nearly-isometric correspondences~\cite{litany2017deep,roufosse2019unsupervised}, maps with bounded distortion~\cite{eisenberger2021neuromorph,eisenberger2020deep} or partial views of the same non-rigid object~\cite{attaiki2021dpfm,rodola2016partial}. Others specialize in distinct classes of shapes like deformable human bodies~\cite{bhatnagar2020loopreg,groueix20183dcoded,marin2018farm}. In contrast, we demonstrate in our experiments that our proposed multi-matching approach excels at a broad range of challenging settings, including non-isometric pairs, poses with topological noise from self-intersections, and inter-class matching.

\begin{figure*}
    \centering
    \includegraphics[width=0.95\linewidth]{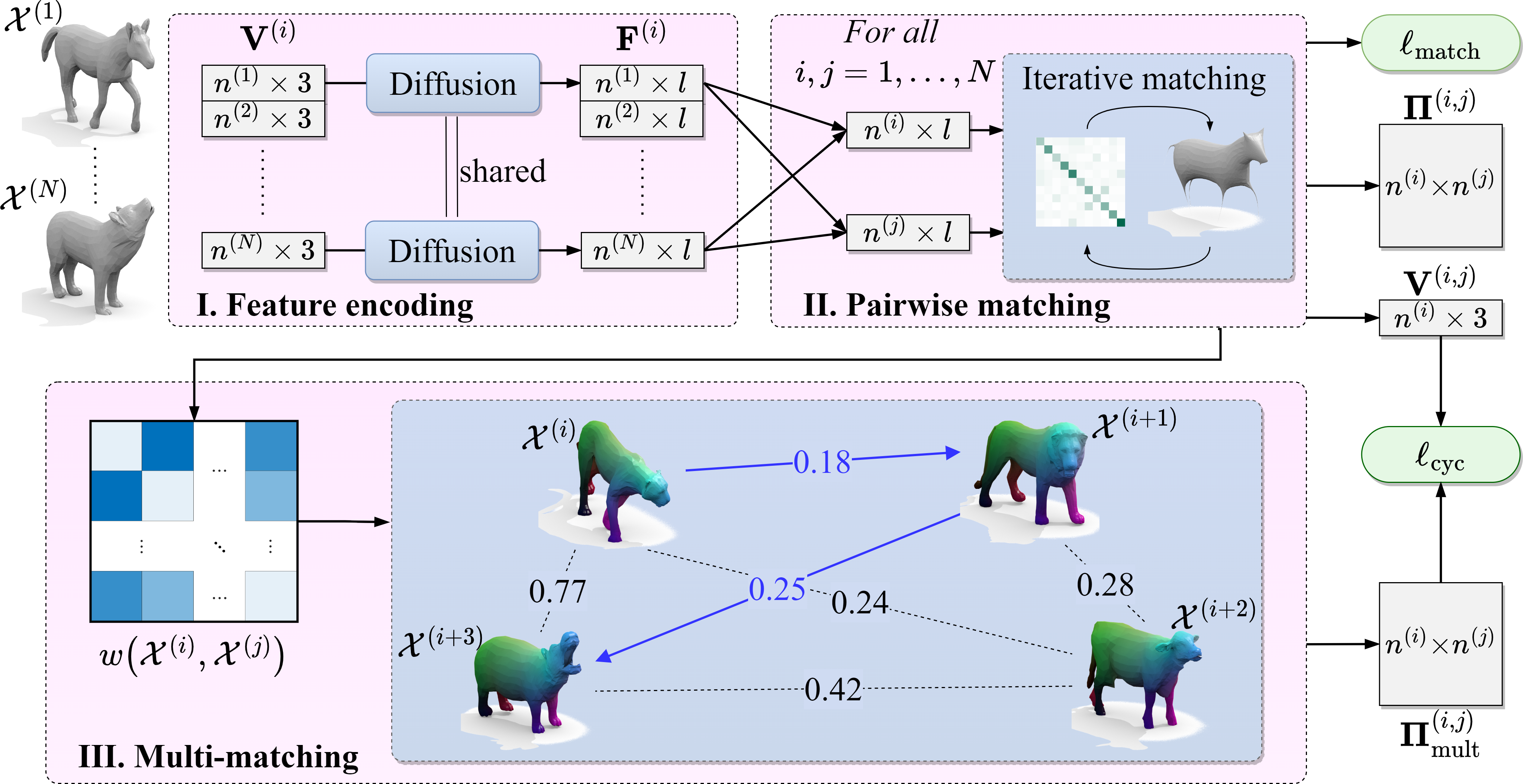}
    \caption{\textbf{Pipeline overview.} For a collection of shapes $\cS=\bigl\{\cX^{(1)},\dots,\cX^{(N)}\bigr\}$, \textbf{I.}~feature embeddings are extracted with DiffusionNet~\cite{sharp2022diffusionnet} and \textbf{II.}~pairwise correspondences $\mPiij$ are predicted via an iterative, differentiable matching layer~\cite{eisenberger2020deep}. \textbf{III.}~The pairwise matches are utilized to construct a shape graph $\cG=\bigl(\cS,w\bigr)$ with affinity edge weights $w\bigl(\cXi,\cXj\bigr)\geq 0$. During training, we minimize the pairwise matching loss $\ell_\mathrm{match}^{(i,j)}$, as well as the cycle consistency loss $\ell_\mathrm{cyc}^{(i,j)}$ between the pairwise registrations $\mVij$ and multi-matches $\mPiijmult$.
    }
    \label{fig:overview}
\end{figure*}

\subsection{Network architecture}\label{sec:architecture}

We now define the neural network architecture that forms the basis of our proposed approach. It consists of three separate components \textbf{I}-\textbf{III}, see also \Cref{fig:overview} for an overview. The first two modules are standard components found in most learning-based shape matching approaches~\cite{attaiki2021dpfm,donati2020deep,eisenberger2020deep,eisenberger2021neuromorph,halimi2019unsupervised,marin2020correspondence,roufosse2019unsupervised,sharma2020weakly}, namely a learnable feature backbone \textbf{I} and a differentiable, pairwise matching layer \textbf{II}. We briefly outline these here and provide additional details in \Cref{app:architecturedetails}. The multi-matching architecture \textbf{III} is introduced in~\Cref{sec:graphmsm}.

\paragraph{Feature extractor}
The first component \textbf{I} of our model is a standard, learnable feature extraction backbone for representation learning, defined as 
\begin{equation}
    \nfeat:\cXi~\mapsto~\mFi\in\bbR^{m\times l}.
\end{equation}
For a given input shape $\cXi=\bigl(\mVi,\mTi\bigr)$, the mapping $\nfeat$ produces an $l$-dimensional feature embedding $\mFi$ per vertex $\mVi\in\bbR^{m\times 3}$. 
While other choices are possible, we base $\nfeat$ on the off-the-shelf feature backbone DiffusionNet~\cite{sharp2022diffusionnet}. This network refines features via intrinsic heat-diffusion operators. Such operators are agnostic to the input discretization, thereby extremely robust to varying mesh resolutions and sampling densities. At the same time, it is computationally lightweight. For more details on the choice of backbone, see \Cref{app:architecturedetails}.

\paragraph{Pairwise matching}
The second component \textbf{II} of our network is a differentiable, multi-scale matching scheme based on the recent pairwise shape matching method DeepShells~\cite{eisenberger2020deep}. The basis of this approach is the energy function
\begin{equation}\label{eq:ot}
    E_\mathrm{match}\bigl(\mF,\mG;\mPisoft\bigr):=\sum_{i'=1}^m\sum_{j'=1}^n\mPisoft_{i',j'}\bigl\|\mF_{i'}-\mG_{j'}\bigr\|_2^2
\end{equation}
which has its roots in the theory of optimal transport. For a given transport plan $\mPisoft\in[0,1]^{m\times n}$, the energy $E_\mathrm{match}$ specifies the distance between the discrete measures associated with two arbitrary $l$-dimensional feature embeddings $\mF=\bigl(\mF_1,\dots,\mF_m\bigr)\in\bbR^{m\times l}$ and $\mG=\bigl(\mG_1,\dots,\mG_n\bigr)\in\bbR^{n\times l}$. 
Taking the minimum over all possible transport plans $\arg\min_\mPisoft E_\mathrm{match}\bigl(\mF,\mG;\mPisoft\bigr)$ results in the Kantorovich formulation of optimal transport~\cite{peyre2019computational,villani2003topics}. Following the approach described in~\cite{eisenberger2020deep}, we obtain a multi-scale shape matching scheme that minimizes \Cref{eq:ot} in an iterative optimization.
For a given pair of shapes $\cXi$ and $\cXj$, this scheme defines a mapping
\begin{equation}\label{eq:nmatch}
    \nmatch:\bigl(\mFi,\mFj\bigl)~\mapsto~\bigl(\mPiij,\mVij,\ell_\mathrm{match}^{(i,j)}\bigr).
\end{equation}
We provide further details on the exact update steps of the optimization scheme $\nmatch$ in \Cref{app:architecturedetails}. 
From a high-level perspective, $\nmatch$ is defined as a deterministic, differentiable function that takes local feature encodings $\mFi\in\bbR^{m\times l}$ and $\mFj\in\bbR^{n\times l}$ as input and predicts a set of correspondences $\mPiij\in\{0,1\}^{m\times n}$. Additionally, $\nmatch$ outputs a deformed embedding $\mVij\in\bbR^{m\times 3}$ of the vertices of $\cXi$. These coordinates specify a registered version of the first input shape $\cXi$ that closely aligns with the pose of the second input shape $\cXj$. The third output $\ell_\mathrm{match}^{(i,j)}>0$ is a training loss signal.

\subsection{Graph-based multi-shape matching}\label{sec:graphmsm}

\paragraph{Shape graph} We now provide details on our multi-shape matching architecture \textbf{III}. To this end, we start by defining an affinity graph 
\begin{equation}
    \cG:=(\cS,w),~~~\text{with}~~~w:\cS\times\cS\to[0,\infty]
\end{equation}
on the set of training shapes $\cS=\bigl\{\cX^{(1)},\dots,\cX^{(N)}\bigr\}$, see \textbf{III} in \Cref{fig:overview} for a visualization. W.l.o.g., we construct $\cG$ as a complete graph (i.e. undirected, fully connected), where a missing edge between $\cXi$ and $\cXj$ can be specified equivalently by setting the corresponding edge weight to $w(\cXi,\cXj)=\infty$. 

We define the pairwise edge weights $w(\cXi,\cXj)\in[0,\infty]$ such that they represent affinity scores between pairs of shapes $\cXi$ and $\cXj$. By convention, small values $w(\cXi,\cXj)\approx 0$ reflect that $\cXi$ and $\cXj$ have a comparably similar geometric structure. 
Since our method is fully unsupervised, we have no a priori knowledge of such affinities and thereby have to infer them directly from the geometries $\cXi$ and $\cXj$.
To this effect, we propose a simple heuristic for a given pair of shapes $\cXi=\bigl(\mVi,\mTi\bigr)$ and $\cXj=\bigl(\mVj,\mTj\bigr)$ and define the (symmetric) affinity score $w$ as
\begin{multline}\label{eq:weights}
    w\bigl(\cXi,\cXj\bigr):=\min\biggl\{E_\mathrm{match}\bigl(\mVij,\mVj;\mPiij\bigr),\\E_\mathrm{match}\bigl(\mVji,\mVi;\mPiji\bigr)\biggr\}.
\end{multline}
In this context, $E_\mathrm{match}$ is the (self-supervised) matching energy defined in \Cref{eq:ot}, while $\mPiij$ and $\mVij$ are the putative correspondences and registrations produced by~\Cref{eq:nmatch}, respectively. The intuition behind this choice of edge weights $w$ is that a small matching energy $E_\mathrm{match}$ implies a  high correspondence accuracy, which is in turn indicative of a high geometric similarity between the input poses $\cXi$ and $\cXj$.

\paragraph{Multi-matching} 
Since we define the edge weights $w$ according to the self-supervised matching score $E_\mathrm{match}$, small weights $w(\cXi,\cXj)$ generally correlate with a high correspondence accuracy of $\mPiij$. Based on this assumption, we obtain multi-shape matches from the putative correspondences $\mPiij$ via the following expression
\begin{subequations}\label{eq:pirefine}
\begin{multline}\label{eq:pirefinea}
    \bigl(i,s_1,\dots,s_{M-1},j\bigr):=\mathrm{Dijkstra}\bigl(\cXi,\cXj;\cG\bigr)
\end{multline}
\begin{equation}\label{eq:pirefineb}
    \mPiijmult:=\mPi^{(i,s_1)}\circ\mPi^{(s_1,s_2)}\circ\dots\circ\mPi^{(s_{M-1},j)}.
\end{equation}
\end{subequations}
Rather than matching a pair of shapes $\cXi$ and $\cXj$ directly, the multi-shape correspondence maps are passed along shortest paths $\cXi,\cX^{(s_1)},\dots,\cX^{(s_{M-1})},\cXj$ in the graph $\cG$. The approach thereby favors edges with a close affinity, i.e., a small pairwise matching cost $w\bigl(\cX^{(s_{k})},\cX^{(s_{k+1})}\bigr)$.
In our experiments, we demonstrate that this simple heuristic yields significant empirical improvements for a broad range of non-rigid matching tasks. 

In practice, we utilize the multi-matching from~\Cref{eq:pirefine} for two distinct use-cases: For once, we can directly query the improved maps $\mPiijmult$ at test time.
Additionally, we promote cycle-consistency during training via the following loss term
\begin{equation}\label{eq:losscyc}
    \ell_\mathrm{cyc}^{(i,j)}:=E_\mathrm{match}\bigl(\mVij,\mVj;\mPiijmult\bigr).
\end{equation}
This loss $\ell_\mathrm{cyc}^{(i,j)}$ imposes a soft penalty on inconsistencies between the registration $\mVij$ produced by the pairwise matching module $\nmatch$, and the multi-shape correspondences $\mPiijmult$ from~\Cref{eq:pirefine}. As before, $E_\mathrm{match}$ is the matching energy defined in \Cref{eq:ot}.

\subsection{Training protocol}\label{sec:trainingprotocol}
The overall loss function that we minimize during training consists of two individual components
\begin{equation}\label{eq:losstotal}
    \ell:=\mathbb{E}_{\cXi,\cXj\sim\cS}\biggl[\ell_\mathrm{match}^{(i,j)}+\lambda_\mathrm{cyc}\ell_\mathrm{cyc}^{(i,j)}\biggr].
\end{equation}
Our complete pipeline is depicted in \Cref{fig:overview}. The whole network is trained end-to-end.
In each training iteration, the backbone \textbf{I} and pairwise matching module $\textbf{II}$ are queried in sequence to produce a pairwise matching for a pair of shapes $\cXi$ and $\cXj$. The shape graph module \textbf{III} then produces the cycle-consistency loss $\ell_\mathrm{cyc}^{(i,j)}$. The shape graph $\cG$ is updated regularly after a fixed number of epochs, taking into account the pairwise matches $\mPiij$ for all $i,j=1,\dots,N$. For more details on the training schedule and choices of hyperparameters, see~\Cref{app:trainingdetails}. 

\section{Experiments}\label{sec:experiments}

We provide various benchmark evaluations for non-rigid shape matching. We consider classical nearly-isometric datasets in~\Cref{sec:nearlyisometric}, as well as more specialized benchmarks for matching with topological changes in \Cref{sec:topologicalmatching} and inter-class pairs in \Cref{sec:interclassmatching}. In~\Cref{sec:comparisongraphtopologies}, we compare different types of shape graph topologies. In~\Cref{sec:ablationarchitecture}, we provide an ablation study of our model, assessing the significance of individual network components. 

\paragraph{Baselines}
We compare G-MSM to existing deep learning approaches for unsupervised, deformable 3D shape correspondence. To this end, we consider both standard pairwise matching~\cite{halimi2019unsupervised,roufosse2019unsupervised,sharma2020weakly,sharp2022diffusionnet,eisenberger2020deep,eisenberger2021neuromorph} and multi-matching approaches~\cite{cao2022unsupervised,huang2022multiway}. Since there are, to date, only very few learning-based multi-matching approaches, we additionally include Consistent ZoomOut~\cite{huang2020consistent} as a recent axiomatic multi-matching approach.

\paragraph{Evaluation}
For each experimental setting, we report the mean geodesic correspondence error over all pairs of a given test set category. All evaluations are performed in accordance with the standard Princeton benchmark protocol~\cite{kim11}. 

\begin{table}
    \resizebox{\linewidth}{!}{
    \begin{tabular}{l|cccccc}
    \toprule[0.2em]
    Method  & \textbf{FAUST} & \textbf{SCAPE} & \textbf{F} on \textbf{S} & \textbf{S} on \textbf{F} & \textbf{SUR} & \textbf{SH'19} \\
    \toprule[0.2em]
    UFM~\cite{halimi2019unsupervised} & \pz5.7 & 10.0 & 12.0 & \pz9.3 & \pz9.2 & 15.5 \\
    SURFM~\cite{roufosse2019unsupervised} & \pz7.4 & \pz6.1 & 19.0 & 23.0 & 38.9 & 37.7 \\
    WFM~\cite{sharma2020weakly} & \pz1.9 & \pz4.9 & \pz8.0 & \pz4.3 & 38.5 & 15.0 \\
    DiffNet~\cite{sharp2022diffusionnet} & \pz1.9 & \pz2.6 & \pz2.7 & \pz1.9 & \pz8.8 & 11.0 \\
    DS~\cite{eisenberger2020deep} & \pz1.7 & \pz2.5 & \pz5.4 & \pz2.7 & \pz2.7 & 12.1 \\
    NM~\cite{eisenberger2021neuromorph} & \textbf{\pz1.5} & \pz4.0 & \pz6.7 & \pz2.0 & \pz9.7 & \pz2.8 \\\cdashline{1-7}
    CZO~\cite{huang2020consistent} & \pz2.2 & \pz2.5 & \pz\textbf{--} & \pz\textbf{--} & \pz2.2 & \pz6.3 \\
    UDM~\cite{cao2022unsupervised} & \textbf{\pz1.5} & \pz2.0 & \pz3.2 & \pz3.2 & \pz3.1 & 22.8 \\
    SyNoRiM~\cite{huang2022multiway} & \pz7.9 & \pz9.5 & 21.9 & 24.6 & 12.7 & \pz7.5 \\
    Ours w/o \textbf{III} & \pz1.7 & \pz3.3 & \pz4.2 & \pz1.7 & \pz8.1 & \pz6.2 \\
    Ours & \textbf{\pz1.5} & \textbf{\pz1.8} & \textbf{\pz2.1} & \textbf{\pz1.5} & \textbf{\pz2.1} & \textbf{\pz2.7} \\
    \bottomrule[0.1em]
    \end{tabular}
    }
    \caption{\textbf{Nearly isometric matching.} A quantitative comparison on four nearly-isometric human shape benchmarks, FAUST~\cite{Bogo:CVPR:2014}, SCAPE~\cite{anguelov2005scape}, SURREAL~\cite{varol2017learning} and SHREC'19~\cite{melzi2019shrec}. Following prior work~\cite{donati2020deep,sharma2020weakly,sharp2022diffusionnet}, we additionally show generalization results when training on FAUST and testing on SCAPE (F on S), and vice versa. We consider both standard, pairwise baselines~\cite{halimi2019unsupervised,roufosse2019unsupervised,sharma2020weakly,sharp2022diffusionnet,eisenberger2020deep,eisenberger2021neuromorph} and multi-matching approaches~\cite{cao2022unsupervised,huang2020consistent,huang2022multiway}.
    }
    \label{fig:geodesic_faustscapeshrec}
\end{table}

\begin{figure*}
    \begin{minipage}[c]{0.65\textwidth}
    \includegraphics[width=0.4906\linewidth]{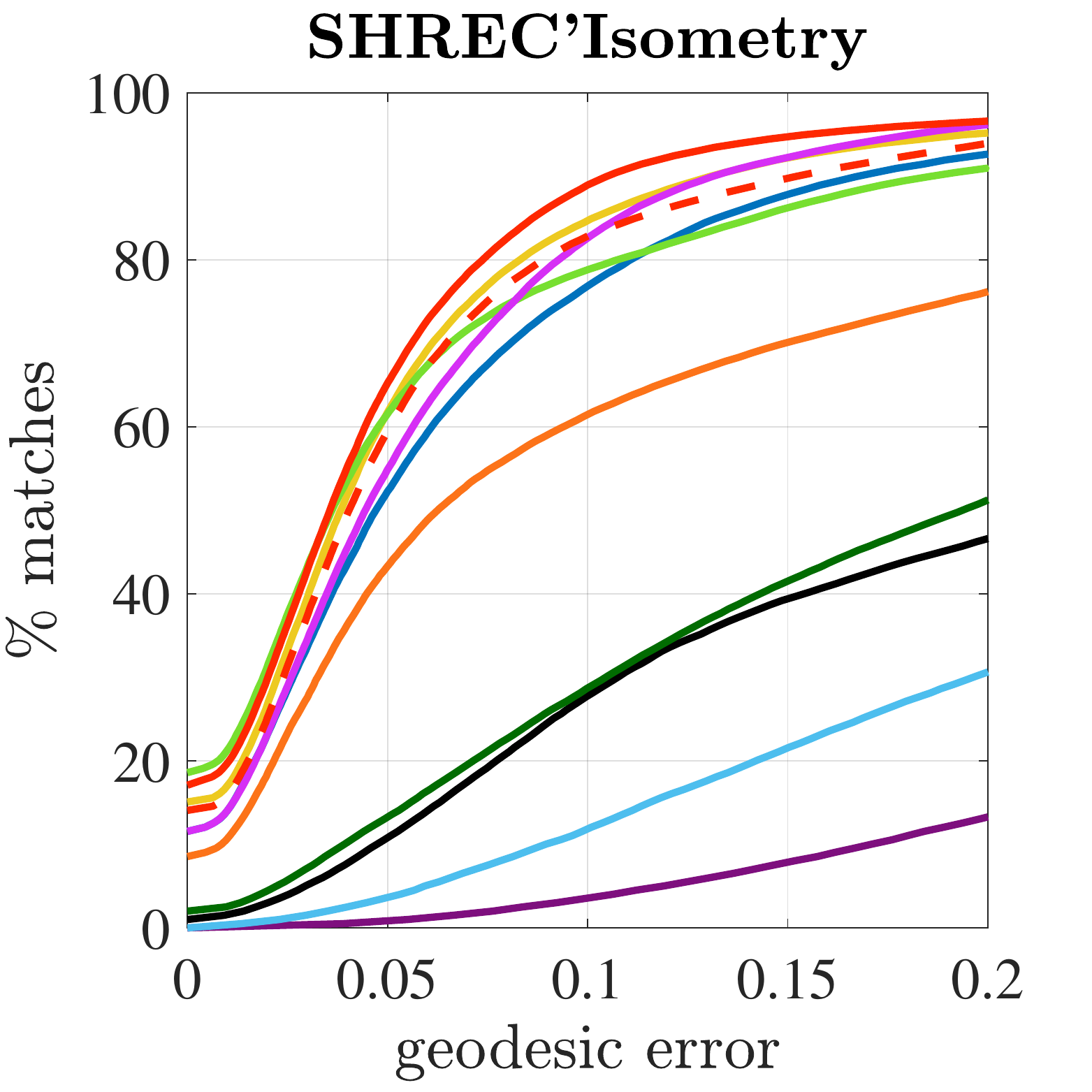}
    \includegraphics[width=0.4688\linewidth]{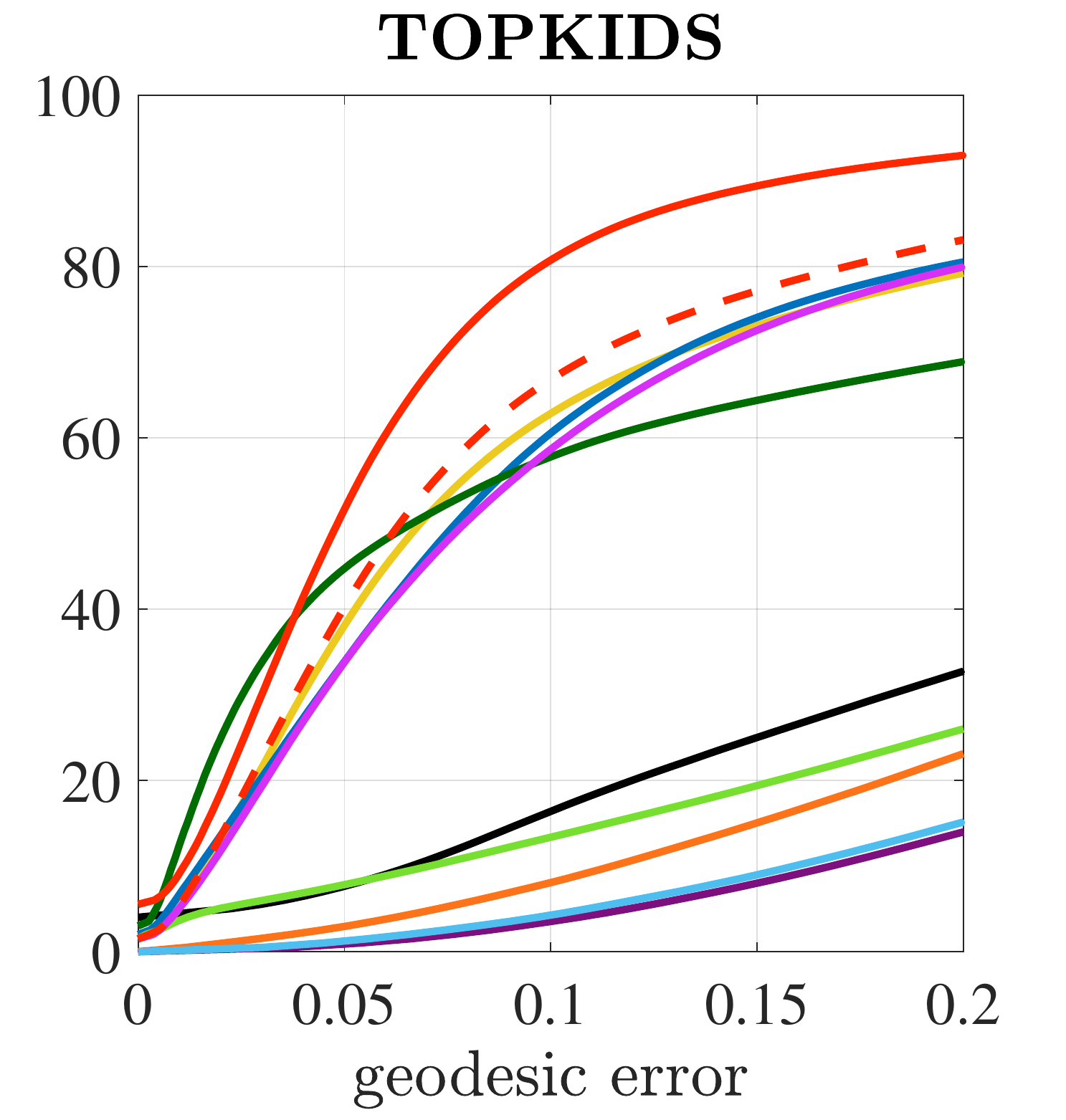}
    \end{minipage}
    \begin{minipage}[c]{0.34\textwidth}
    \resizebox{\linewidth}{!}{
    \begin{tabular}{ccl|cc}
    \toprule[0.2em]
    \hspace{3pt} & \hspace{3pt} & Method  & \textbf{SH'Iso} & \textbf{KIDS} \\
    \toprule[0.2em]
    \cellcolor{UFM}&\cellcolor{UFM} & UFM~\cite{halimi2019unsupervised} & 13.4  & 38.5 \\ 
    \cellcolor{SURFM}&\cellcolor{SURFM} & SURFM~\cite{roufosse2019unsupervised} & 45.6 & 48.6 \\ 
    \cellcolor{WFM}&\cellcolor{WFM} & WFM~\cite{sharma2020weakly} & 38.0 & 47.9  \\ 
    \cellcolor{DiffNet}&\cellcolor{DiffNet} & DiffNet~\cite{sharp2022diffusionnet} & 26.5 & 35.7  \\ 
    \cellcolor{DS}&\cellcolor{DS} & DS~\cite{eisenberger2020deep} & \pz6.3 & 13.7  \\ 
    \cellcolor{NM}&\cellcolor{NM} & NM~\cite{eisenberger2021neuromorph} & \pz7.7 & 13.8  \\\cdashline{1-5} 
    \cellcolor{CZO}&\cellcolor{CZO} & CZO~\cite{huang2020consistent} & \pz7.6 & 39.3  \\ 
    \cellcolor{UDM}&\cellcolor{UDM} & UDM~\cite{cao2022unsupervised} & 23.6 & 18.2  \\ 
    \cellcolor{SyNoRiM}&\cellcolor{SyNoRiM} & SyNoRiM~\cite{huang2022multiway} & \pz6.2 & 13.8  \\ 
    &\cellcolor{Ours} & Ours w/o \textbf{III} & \pz6.3 & 12.0 \\
    \cellcolor{Ours}&\cellcolor{Ours} & Ours & \textbf{\pz5.2} & \textbf{\pz7.9}  \\ 
    \bottomrule[0.1em]
    \end{tabular}
    }
    \end{minipage}
    \caption{\textbf{Matching with topological noise.} A summary of our quantitative comparisons on the topology benchmarks SHREC'Isometry~\cite{dyke2019shrec} and TOPKIDS~\cite{laehner2016shrec}. For both benchmarks, we show the cumulative error curves of our approach (red) and all considered baselines. Additionally, we provide the mean geodesic errors, averaged over all pairs of shapes, respectively (table, right).}
    \label{fig:geodesic_shrec_top}
\end{figure*}

\begin{figure*}
    \vspace{10pt}
    \centering
    \begin{overpic}
    [width=0.9\linewidth]{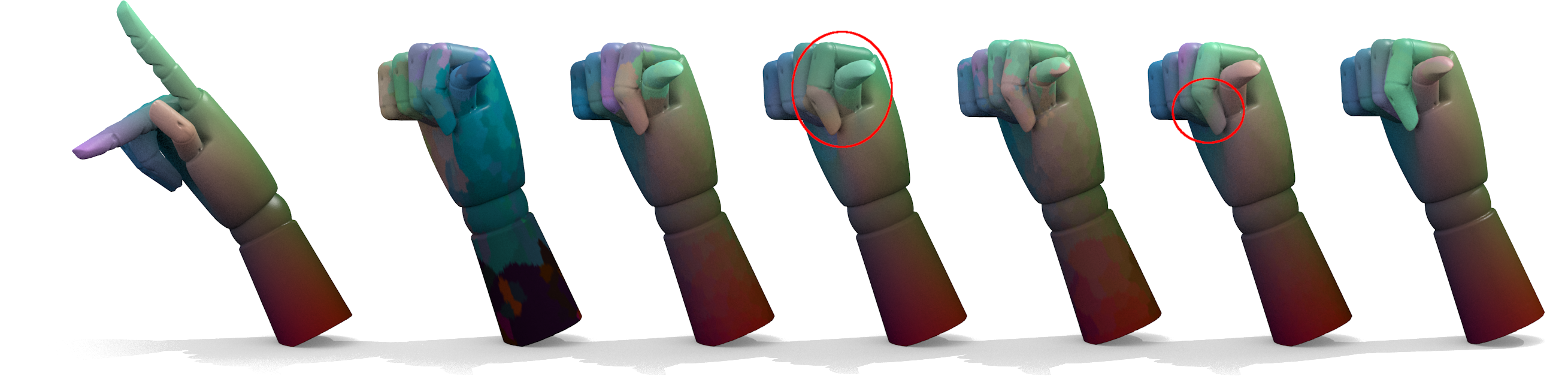}
    \put(6,26){Target}
    \put(24,25){UFM~\cite{halimi2019unsupervised}}
    \put(36,25){CZO~\cite{huang2020consistent}}
    \put(49.5,25){DS~\cite{eisenberger2020deep}}
    \put(62,25){NM~\cite{eisenberger2021neuromorph}}
    \put(72,25){SyNoRiM~\cite{huang2022multiway}}
    \put(88,25){Ours}
    \end{overpic}
    \vspace{-5pt}
    \caption{\textbf{Qualitative baseline comparison.} We consider a pair of real 3D scan meshes from SHREC'Iso~\cite{dyke2019shrec}, corresponding to the quantitative experiments shown in~\Cref{fig:geodesic_shrec_top}. The two geometries are subject to topological merging due to self-contact of different fingers and parts of the hand. Correspondences are shown via a colormap for our method, as well as several baseline approaches. For this example, the best results are achieved by~\cite{eisenberger2020deep},~\cite{huang2022multiway} and our method. However, for both~\cite{eisenberger2020deep} and~\cite{huang2022multiway}, the front part of the index and middle fingers are erroneous (tip of index finger should be bright green). See our supplementary material for additional qualitative examples.
    }
    \label{fig:qual_shrec19}
\end{figure*}
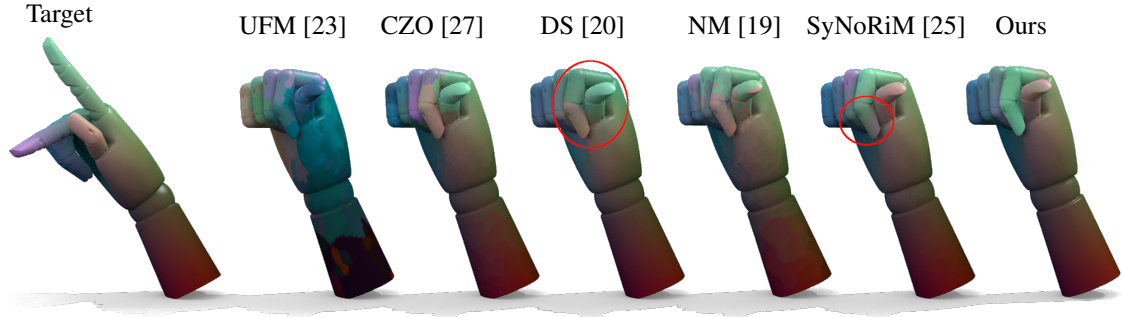

\subsection{Nearly isometric matching}\label{sec:nearlyisometric}
\paragraph{Datasets} 
We evaluate our method on four classical, nearly-isometric datasets.
FAUST~\cite{Bogo:CVPR:2014} contains 10 humans in 10 different poses each and SCAPE~\cite{anguelov2005scape} contains 71 diverse poses of the same individual. We follow the standard benchmark protocol from existing work~\cite{donati2020deep,sharma2020weakly,sharp2022diffusionnet}. Specifically, we consider the more challenging remeshed geometries from \cite{ren2018orientation} to avoid overfitting to a particular triangulation. 
SURREAL~\cite{varol2017learning} consists of synthetic SMPL~\cite{SMPL:2015} meshes fit to raw 3D motion capture data.
The last benchmark, which is the most challenging among the four, is SHREC'19 Connectivity~\cite{melzi2019shrec}. It contains human shapes in different poses with significantly varying sampling density and quality, as well as a small number of non-isometric poses.

\paragraph{Discussion}
The results on these four benchmarks are summarized in~\Cref{fig:geodesic_faustscapeshrec}. Our method obtains state-of-the-art performance in all considered settings. Remarkably, these results were achieved directly through querying our network, whereas many baselines require correspondence postprocessing~\cite{cao2022unsupervised,eisenberger2021neuromorph,halimi2019unsupervised,roufosse2019unsupervised,sharma2020weakly,sharp2022diffusionnet}. Furthermore, the results underline that the shape graph module \textbf{III} plays a critical role in our pipeline for optimal performance.

\begin{figure*}
\begin{minipage}[c]{0.57\textwidth}
\resizebox{\linewidth}{!}{
\begin{tabular}{l|cccccccc}
\toprule[0.2em]
 & & \textbf{SH'20} on & \multicolumn{6}{c}{\textbf{SH'20} on \textbf{TOSCA}} \\ \cline{4-9}
 & \textbf{SH'20} &  \textbf{SMAL} & \small Cat & \small Centaur & \small Dog & \small Horse & \small Human & \small Wolf \\
\toprule[0.2em]
UFM~\cite{halimi2019unsupervised} & 39.8 & 32.9 & 39.4 & 39.2 & 37.5 & 34.1 & 49.6 & \pz4.4 \\
SURFM~\cite{roufosse2019unsupervised} & 53.4 & 37.7 & 54.0 & 57.7 & 57.9 & 57.0 & 65.8 & 55.3 \\
WFM~\cite{sharma2020weakly} & 31.4 & 20.2 & 20.6 & 21.9 & 16.7 & 22.4 & 38.1 & \pz5.7 \\
DiffNet~\cite{sharp2022diffusionnet} & 40.5 & 18.2 & 14.2 & \pz8.3 & 13.6 & \pz9.1 & 24.5 & \pz2.6 \\
DS~\cite{eisenberger2020deep} & 35.0 & 10.8 & \pz7.6 & \pz9.1 & \pz5.5 & \pz2.5 & 10.1 & \pz2.1 \\
NM~\cite{eisenberger2021neuromorph} & \textbf{10.0} & \pz9.9 & 16.8 & 12.7 & 14.6 & 11.2 & 29.7 & \pz1.5 \\\cdashline{1-9}
CZO~\cite{huang2020consistent} & 21.7 & \pz\textbf{--} & \pz\textbf{--} & \pz\textbf{--} & \pz\textbf{--} & \pz\textbf{--} & \pz\textbf{--} & \pz\textbf{--} \\
UDM~\cite{cao2022unsupervised} & 52.6 & 25.5 & 40.7 & 34.3 & 43.6 & 43.0 & 45.8 & 34.3 \\
SyNoRiM~\cite{huang2022multiway} & 10.4 & \pz5.7 & 12.8 & 11.6 & 10.6 & \pz7.1 & 28.2 & \pz2.0 \\
Ours w/o \textbf{III} & 11.1 & \pz3.4 & \pz6.3 & \pz6.0 & \pz4.9 & \pz2.6 & 20.1 & \pz2.2 \\
Ours & 10.6 & \textbf{\pz2.6} & \textbf{\pz5.2} & \textbf{\pz2.0} & \textbf{\pz3.0} & \textbf{\pz2.2} & \textbf{\pz8.3} & \textbf{\pz1.4} \\
\bottomrule[0.1em]
\end{tabular}
}
\vspace{5pt}
\end{minipage}
\begin{minipage}[c]{0.42\textwidth}
\includegraphics[width=\linewidth]{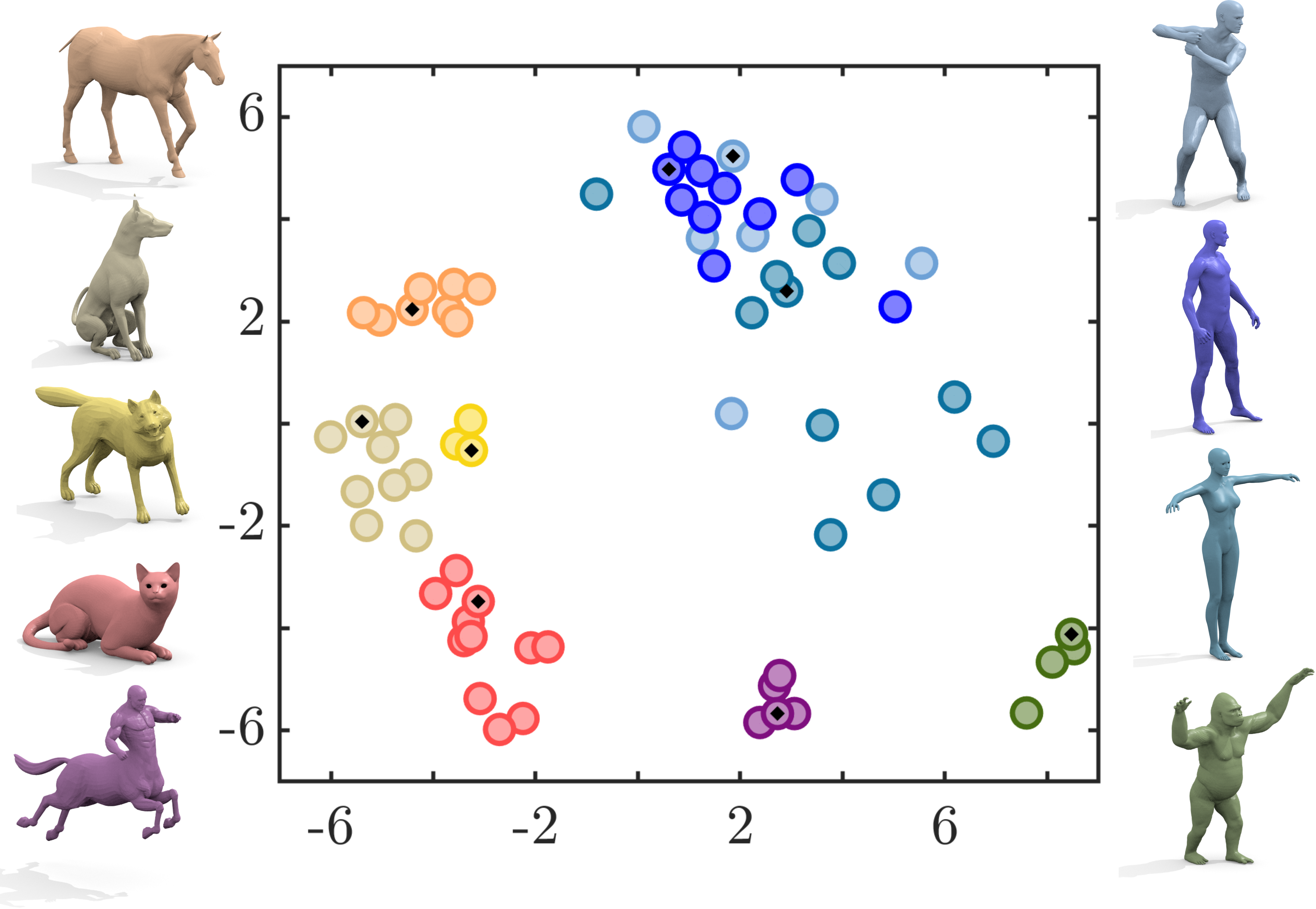}
\end{minipage}
\centering
\caption{\textbf{Inter-class matching.} (left) A comparison of our approach to the considered baselines on SHREC'20~\cite{dyke2020shrec}. We further assess the generalization from training on SHREC'20 to two different synthetic datasets, SMAL~\cite{zuffi20173d} and TOSCA~\cite{bronstein2008numerical}.
(right) Additionally, we visualize the shape graph node embeddings of our approach on TOSCA~\cite{bronstein2008numerical} through 2D multi-dimensional scaling. Since the learned edge weights express affinity scores, shapes with similar geometries tend to cluster together. The three human classes (shades of blue) form one big cluster, which indicates that the intra-category pose variety is higher than geometric differences between subjects. Shapes with four legs (orange $\hateq$ horse, red $\hateq$ cat, \dots) and two legs (blue $\hateq$ human, green $\hateq$ gorilla) are linearly separable in the 2D MDS space. Interestingly, the centaur classes' embedding (purple) lies exactly between these two categories.
}
\label{fig:interclass}
\end{figure*}

\subsection{Matching with topological changes}\label{sec:topologicalmatching}
\paragraph{Datasets} 
The benchmarks SHREC'Isometry~\cite{dyke2019shrec} and TOPKIDS~\cite{laehner2016shrec} focus on matching with topological noise. This is a common phenomenon when working with real scans, where the mesh topology is often corrupted by self-contact of separate parts of the scanned objects. Such topological merging severely affects the correspondence estimation since it distorts the intrinsic shape geometry non-isometrically. 
The first benchmark SHREC'Isometry~\cite{dyke2019shrec} contains real scans of different humanoid puppets and hand models. A majority of poses in their `heteromorphic' test set are subject to topological changes, see also \Cref{fig:qual_shrec19} for an example. The TOPKIDS~\cite{laehner2016shrec} dataset contains synthetic shapes of human children where topological merging is emulated by computing the outer hull of intersecting geometries, see~\Cref{fig:teas} for an example.

\paragraph{Discussion}
Quantitative results are shown in~\Cref{fig:geodesic_shrec_top}.
We observe that topological merging commonly leads to unstable behavior for methods that rely on intrinsic priors like preservation of the Laplace-Beltrami operator~\cite{cao2022unsupervised,huang2020consistent,roufosse2019unsupervised,sharma2020weakly,sharp2022diffusionnet} or pairwise geodesic distances~\cite{halimi2019unsupervised}. It further inhibits approaches that learn to morph input geometries~\cite{eisenberger2021neuromorph,huang2022multiway} with explicit deformation priors, since merged regions tend to adhere to each other, see, e.g., the discussion on failure cases in~\cite[Sec. 7]{huang2022multiway}. 
Our pipeline decreases the correspondence error by a decisive margin of $19\%$ for SHREC'Iso and $73\%$ for TOPKIDS. We provide a qualitative comparison in~\Cref{fig:qual_shrec19}, as well as additional examples in~\Cref{app:qualitative}.

\subsection{Inter-class matching}\label{sec:interclassmatching}

\paragraph{Datasets} 
The SHREC'20~\cite{dyke2020shrec} challenge contains real scans of various four-legged animal models, including: elephant, giraffe, bear, and many more. These geometries were obtained from inhomogeneous acquisition sources, i.e., different types of scanners and 3D reconstruction pipelines. 
Sparse ground-truth correspondences were obtained through manual annotation. We further assess the generalization to additional test sets from the synthetic SMAL~\cite{zuffi20173d} and TOSCA~\cite{bronstein2008numerical} datasets. SMAL contains inter-class pairs between different animal classes, whereas TOSCA contains nearly-isometric pairs with both animal and human classes. 

\paragraph{Discussion}
The resulting accuracies are summarized in \Cref{fig:interclass}. 
Our approach yields the most stable results overall. 
Several baselines suffer from unstable behavior for animals from SHREC'20, because they depend on either noisy SHOT~\cite{tombari2010SHOT} input features~\cite{eisenberger2020deep}, intrinsic priors that favor near isometries~\cite{huang2020consistent,sharma2020weakly,sharp2022diffusionnet}, or both~\cite{cao2022unsupervised,halimi2019unsupervised,roufosse2019unsupervised}. While methods with an explicit deformation prior~\cite{eisenberger2021neuromorph,huang2022multiway} perform well on SHREC'20, they do not generalize well to unseen test poses from SMAL and TOSCA.
Our method learns a topology-aware shape graph prior and thereby gets the best out of both worlds, i.e., robustness to inter-class pairs and strong generalization to unseen test pairs.

\begin{table*}

\resizebox{!}{42pt}{
\begin{tabular}{ll|ccccc}
\toprule[0.2em]
\multicolumn{2}{c}{\multirow{2}{*}{\textbf{SH'Iso}}} & \multicolumn{5}{c}{\textit{Train on}} \\
\multicolumn{2}{c}{} & Full & MST & TSP & Star & w/o \\
\toprule[0.2em]
\multirow{5}{*}{\rotatebox[origin=c]{90}{\textit{Test on}}} 
& Full     & \yc5.16 & \textbf{5.09} & 5.20 & 5.90 & 5.17\\
& MST      & 5.68 & \yc5.49 & 5.50 & 6.36 & 5.49\\
& TSP     & 6.08 & 5.62 &\yc 5.80 & 6.94 & 6.51\\
& Star     & 5.54 & 5.26 & 5.44 & \yc6.71 & 6.02\\
& w/o      & 5.32 & 5.27 & 5.42 & 6.33 & \yc6.27\\
\bottomrule[0.1em]
\end{tabular}
}
\hspace{0.05\linewidth}
\resizebox{!}{42pt}{
\begin{tabular}{ll|ccccc}
\toprule[0.2em]
\multicolumn{2}{c}{\multirow{2}{*}{\textbf{TOPKIDS}}} & \multicolumn{5}{c}{\textit{Train on}} \\
\multicolumn{2}{c}{} & Full & MST & TSP & Star & w/o \\
\toprule[0.2em]
\multirow{5}{*}{\rotatebox[origin=c]{90}{\textit{Test on}}} 
& Full     & \yc\textbf{\pz7.92} & \pz8.13 & \pz8.44 & 11.03 & \pz9.13 \\
& MST      & \pz8.56 & \yc\pz8.62 & \pz9.39 & 10.57 & \pz9.98 \\
& TSP    & 13.18 & 12.33 & \yc13.10 & 19.72  & 15.07\\
& Star     & \pz8.61 & \pz8.84 & \pz8.34 & \yc11.92  & \pz9.63 \\
& w/o & 10.62 & 10.64 & 11.61 & 13.62 & \yc12.02\\
\bottomrule[0.1em]
\end{tabular}
}
\vspace{5pt}
\centering
    \caption{\textbf{Graph topology comparison.} We compare the quantitative performance of our model for different graph topologies $\cG$. Specifically, we revisit the experiment from~\Cref{fig:geodesic_shrec_top} and report the mean geodesic error on SHREC'Iso~\cite{dyke2019shrec} and TOPKIDS~\cite{laehner2016shrec}. The standard `full' graph is compared to three sparse topologies `MST', `TSP', `star' graph, as well as the `w/o \textbf{III}' variant of our pipeline.
    }
    \label{tab:ablationgraphtype}

\end{table*}

\subsection{Sparse graph topologies}\label{sec:comparisongraphtopologies}

Throughout our experiments, we use complete shape graphs $\cG$ with a full set of $\frac{N(N-1)}{2}$ edges, as specified in~\Cref{sec:graphmsm}. Here, we explore a few alternative graph topologies with sparse connectivity patterns, i.e., $\mathcal{O}(N)$ edges: (ii) Minimal spanning trees (MST), (iii) minimal paths solving the traveling salesman problem (TSP) and (iv) star graphs, where all nodes are connected to one center node. For a detailed discussion and visualizations of these graph types, see~\Cref{app:graphtopologies}. We compare (ii)-(iv) to (i) the standard full graph and (v) our pipeline without the graph module \textbf{III}. A given graph type can be employed either for the cycle consistency loss in~\Cref{eq:losscyc} during training or for map refinement at test time. For a complete picture, we report results for all $5\times 5=25$ possible combinations of training/evaluation graph types. 

Our results in~\Cref{tab:ablationgraphtype} indicate that, while the full graph is generally the most accurate, the sparse topologies often perform comparably, especially MST. This makes them a viable alternative to the full graph in certain scenarios with limited resources, both in terms of the required memory and query time. We provide a comprehensive cost analysis in~\Cref{app:computationalcomplexity}.
In most cases, using the shape graph during training is beneficial, even when no graph is available at test time (\Cref{tab:ablationgraphtype}, bottom row). This makes it relevant for online applications where not all test pairs are available at once.
Regardless of the graph type, it is generally preferable to include some version of our graph module \textbf{III} rather than directly using the pairwise correspondences `w/o'.

\begin{table}
\resizebox{0.9\linewidth}{!}{
\begin{tabular}{ll|cccc}
\toprule[0.2em]
& \multicolumn{1}{c}{} & \multicolumn{2}{c}{\textbf{TOPKIDS}} & \multicolumn{2}{c}{\textbf{SHREC'20}} \\
&\multicolumn{1}{r}{\emph{with} \textbf{III}}& \cmark & \xmark & \cmark & \xmark\\
\toprule[0.05em]
\multirow{5}{*}{\rotatebox[origin=c]{90}{\small Feature \textbf{I}}} 
& \textbf{I.a}~SpecConv~\cite{eisenberger2020deep} & \pz8.53 & 13.68 & 28.54 & 34.99\\
& \textbf{I.b}~SHOT~\cite{tombari2010SHOT} & \pz7.93 & 13.66 & 22.74 & 29.81 \\
& \textbf{I.c}~ResNet~\cite{litany2017deep} & \pz7.94 & 13.14 & 39.69 & 40.66 \\
& \textbf{I.d}~PointNet~\cite{qi2017pointnet} & \pz8.78 & 14.10 & 11.01 & 11.54 \\
& \textbf{I.e}~GraphNN~\cite{eisenberger2021neuromorph} & 14.18 & 25.57 & 14.53 & 18.33 \\
\midrule[0.05em]
\multirow{3}{*}{\rotatebox[origin=c]{90}{\small Match \textbf{II}}} 
& \textbf{II.a}~FM~\cite{ovsjanikov2012functional} & 39.12 & 40.66 & 50.58 & 51.37 \\
& \textbf{II.b}~Sinkhorn~\cite{cuturi2013sinkhorn} & 12.25 & 14.81 & 11.58 & 12.66 \\
& \textbf{II.c}~Softmax~\cite{eisenberger2021neuromorph} & 12.78 & 13.46 & 11.49 & 13.47 \\
\midrule[0.05em]
& G-MSM (ours) & \textbf{\pz7.92} & \textbf{12.02} & \textbf{10.65} & \textbf{11.06} \\
\bottomrule[0.1em]
\end{tabular}
}
\vspace{5pt}
\centering
\caption{\textbf{Ablation network architecture.} We compare several off-the-shelf network architectures for the feature backbone \textbf{I} and matching module \textbf{II} to our full model, as defined in~\Cref{sec:architecture}. For each setting, we contrast the results obtained with (\cmark) and without (\xmark) the graph-based multi-matching module \textbf{III}.
}
\label{tab:ablationarchitecture}

\end{table}

\subsection{Ablation study}\label{sec:ablationarchitecture}

Our proposed architecture consists of several basic building blocks \textbf{I}-\textbf{III}, as defined in~\Cref{sec:architecture} and~\Cref{sec:graphmsm}. While the shape-graph module \textbf{III} is unique to our approach, the feature backbone \textbf{I} and matching module \textbf{II} can, in principle, be replaced by any analogous off-the-shelf architectures.
We compare several popular alternatives: For the feature backbone, we consider {I.a} the spectral convolution architecture from~\cite{eisenberger2020deep}, {I.b} our network with SHOT~\cite{tombari2010SHOT} input features, {I.c} the ResNet architecture from~\cite{litany2017deep}, {I.d} PointNet~\cite{qi2017pointnet}, as well as {I.e} the message passing architecture from~\cite{eisenberger2021neuromorph}. For the differentiable matching module, we compare {II.a} a functional map layer~\cite{ovsjanikov2012functional}, {II.b} a single Sinkhorn layer~\cite{cuturi2013sinkhorn} and {II.c} a standard per-point Softmax~\cite{eisenberger2021neuromorph}.

We then replace either the feature backbone {I.a}-{I.e} or matching module {II.a}-{II.c} in our method and observe how it affects the accuracy on TOPKIDS~\cite{laehner2016shrec} and SHREC'20~\cite{dyke2020shrec} from~\Cref{sec:topologicalmatching} and~\Cref{sec:interclassmatching}, respectively. The results are summarized in~\Cref{tab:ablationarchitecture}. Replacing either module \textbf{I} or \textbf{II} in our approach leads to a drop in performance. Moreover, we see that, regardless of the concrete architecture, our multi-matching approach \textbf{III} (\cmark in~\Cref{tab:ablationarchitecture}) improves the performance over the pairwise matches (\xmark).

\section{Conclusion}
We propose G-MSM, a novel multi-matching approach for non-rigid shape correspondence. 
For a given collection of 3D meshes, we define a shape graph $\cG$ which approximates the underlying shape data manifold. Its edge weights $w$ are extracted from putative pairwise correspondence signals in a self-supervised manner. Our network promotes cycle-consistency of optimal paths in $\cG$. Thus, it produces context-aware multi-matches that are informed by commonalities and salient geometric features across all training poses. In our experiments, we demonstrate that this simple strategy yields significant improvements in correspondence accuracy on a wide range of challenging, real-world 3D mesh benchmarks.

\paragraph{Limitations \& future work}
Our method can effectively learn the underlying canonical shape topology from a collection of 3D meshes. On the other hand, it naturally relies on at least some of the poses to convey this latent topology. In the extreme case, if, instead of a shape collection, we only have $N=2$ poses available, our multi-shape pipeline does not yield an improvement over the naive pairwise maps. 

Since our affinity weights $w$ are constructed through a self-supervised heuristic, it is difficult to provide theoretical guarantees that the multi-matching is, without fail, always superior to the putative pairwise correspondences. However, our model has an in-built robustness to this type of error: If the pairwise putative correspondences are very accurate, its corresponding edge weights are low and the network tends to favor the direct connections. 
Empirically, we observe that the multi-matching $\textbf{III}$ consistently improves the accuracy over the putative matches.

Another potential direction for future research is extending our framework to allow for partial views, e.g., by leveraging recent advances on learnable partial functional maps~\cite{attaiki2021dpfm}.

\paragraph{Societal impact}
Advancing the robustness and accuracy of shape correspondence methods has the potential to open up new avenues for future applications based on 3D scan data. Our algorithm constitutes one small advancement in this effort of extending computer vision algorithms to the 3D domain. Since our algorithm is fully unsupervised, it can directly reduce deployment costs as no manual correspondence annotations are required to train our model.
Shape correspondence is a fundamental building block at the heart of many 3D vision algorithms and we do not anticipate any immediate risk of misuse associated with this work. 

\section*{Acknowledgements}
We acknowledge support by the ERC Advanced Grant SIMULACRON and the Munich School for Data Science.

{\small
\bibliographystyle{ieee_fullname}
\bibliography{ms}
}

\clearpage

\appendix

\section{Implementation details}\label{app:implementationdetails}

\subsection{Training details}\label{app:trainingdetails}

In the following, we provide additional details on our training protocol and choice of parameters. Throughout our experiments, our model was trained on a single NVIDIA Quadro RTX 8000 graphics card with 48GB VRAM.

\paragraph{Data preprocessing}
For a given shape collection $\cS$, we apply a few data standardization steps to ensure training stability. For each $\cXi$, we normalize the scale of the shape by setting the approximate geodesic diameter to a constant value $\sqrt{\mathrm{area}(\cXi)}=\frac{2}{3}$. The pose is further centered around the origin by setting the mean vertex position to $0\in\bbR^3$. The eigenvalues and eigenvectors required for our method are precomputed prior to training our model. Otherwise, our method is directly applicable to any collection of shapes that fulfill the weak pose alignment as discussed in~\Cref{sec:problemformulation}.

\paragraph{Training scheduling}
Our general training protocol is outlined in~\Cref{sec:trainingprotocol} and illustrated visually in~\Cref{fig:overview} of the main paper. 

In each forward pass, we first query the DiffusionNet backbone $\nfeat$ to obtain sets of local features $\mFi:=\nfeat\bigl(\cXi\bigr)$ and $\mFj:=\nfeat\bigl(\cXj\bigr)$ for a pair of input shapes $\cXi$ and $\cXj$. In the second step, the matching module $\bigl(\mPiij,\mVij,\ell_\mathrm{match}^{(i,j)}\bigr):=\nmatch\bigl(\mFi,\mFj\bigl)$ computes a set of putative correspondences $\mPiij$, registered vertices $\mVij$ and the corresponding matching loss $\ell_\mathrm{match}^{(i,j)}$. Finally, \Cref{eq:pirefine} produces the multi-shape correspondences $\mPiijmult$, which then allows us to compute the loss $\ell_\mathrm{cyc}^{(i,j)}$ through~\Cref{eq:losscyc}. 

As stated in~\Cref{sec:trainingprotocol}, the shape graph $\cG$ is updated regularly after a fixed number of epochs. This interval is chosen in dependence of the number of training shapes as a round figure that results in around $10k-15k$ training iterations per update. To reduce the computational load, the pairwise correspondences between all pairs of training shapes $\mPiijmult$ are precomputed and stored each time the shape graph is constructed. Additionally, we wait for $5$ shape graph update cycles before activating the cycle-consistency loss. This burn-in period allows the feature extractor and putative correspondence modules to converge to a certain degree which facilitates a stable training and reduces stochasticity. 

\paragraph{Learning}
Our model is trained in an end-to-end manner with the Adam optimizer~\cite{kingma2014adam}, using standard parameters. All the learnable weights are contained in the DiffusionNet backbone $\textbf{I}$. The DeepShells pairwise matching module $\textbf{II}$ and multi-matching module $\textbf{III}$ convert the learned features into correspondences, but these maps themselves are fully deterministic. The backward pass updates the DiffusionNet weights both in terms of the pairwise alignment loss $\ell_\mathrm{match}^{(i,j)}$ and the cycle-consistency loss $\ell_\mathrm{cyc}^{(i,j)}$. The former stems from~\Cref{eq:nmatch} and the latter term is defined in \Cref{eq:losscyc}. The cycle-consistent correspondences themselves $\mPiijmult$ are obtained from non-differentiable operations, since they are the quantized outputs from DeepShells concatenated via Dijkstra's algorithm in~\Cref{eq:pirefinea}. Instead, the gradients from $\ell_\mathrm{cyc}^{(i,j)}$ pass information back to the DeepShells layer $\textbf{II}$ through the registrations $\mVij$. 

\paragraph{Hyperparameters}
We set the cycle-consistency loss weight to $\lambda_\mathrm{cyc}=0.5$. The number of latent dimensions of the DiffusionNet encodings is chosen as $l=128$ and the architecture comprises $4$ consecutive DiffusionNet blocks. For the DeepShells matching layer, we directly use the hyperparameters specified by the original publication and the corresponding source code~\cite{eisenberger2020deep}. Specifically, the number of eigenfunctions used to compute the smooth shell product space poses (see~\Cref{sec:appendixmatchingmodule}) is upsampled on a log-scale between $k_\mathrm{min}=6$ and $k_\mathrm{max}=21$.

\subsection{Architecture details}\label{app:architecturedetails}

\subsubsection{Feature backbone}
Our network proposed in~\Cref{sec:architecture} leverages the recent DiffusionNet~\cite{sharp2022diffusionnet} backbone for local feature extraction. We outline the basic architecture of this module here and refer to~\cite[Sec. 3]{sharp2022diffusionnet} for further technical details.

The core motivation is to model feature propagation of signals $f(x,t)$ on the surface of a shape $\cXi$ as heat diffusion, governed by the standard heat equation
\begin{equation}\label{eq:heateq}
    \frac{\partial}{\partial t}f(x,t)=\Delta f(x,t).
\end{equation}
In this context, $\Delta$ is the intrinsic Laplace-Beltrami operator on the surface $\cXi$. 
In the discrete case, a common approximation is the cotangent Laplacian $\mL:=-\mM^{-1}\mS\in\bbR^{m\times m}$, where $\mM$ and $\mS$ are the mass matrix and stiffness matrix, respectively. 
We further consider the truncated basis of eigenfunctions $\mPsi\in\bbR^{m\times k}$ and corresponding diagonal matrix of eigenvalues $\mLambda\in\bbR^{k\times k}$ of the discretized Laplacian.
For a given signal $\vf\in\bbR^{m}$ on $\cXi$, the heat propagation for a time-interval $t>0$ according to~\Cref{eq:heateq} then results in the approximate solution
\begin{equation}
    H_t(\vf):=\mPsi\exp(t\mLambda)\mPsi^\dagger \vf.
\end{equation}
For a given feature matrix $\mF\in\bbR^{m\times l}$, an individual operator $H_t$ is applied separately to each channel with different learnable time step weights $t$. A key benefit of such propagation operators is that they are indifferent to the sampling density and therefore robust to remeshing and local noise. On the other hand, pure heat diffusion is spatially isotropic and therefore not sufficiently expressive. To break radial symmetry, DiffusionNet additionally leverages a gradient-based feature refinement layer. At every point on the surface, it computes inner products between spatial gradients of the (scalar) feature signals on the tangent plane. 

Putting everything together, an individual DiffusionNet block takes a set of features $\mF$, propagates information both via a spatial diffusion and a spatial gradient layer, and feeds them to a per-point multilayer perceptron (MLP). The first layer is initialized by the input features, defined as the vertex coordinates $\mVi$. For further technical details, we refer the interested reader to the original publication~\cite{sharp2022diffusionnet}.

\subsubsection{Hierarchical pairwise matching}\label{sec:appendixmatchingmodule}
In~\Cref{sec:architecture}, we further introduced our differentiable matching layer $\nmatch$ based on DeepShells~\cite{eisenberger2020deep}. The final map is fully specified by~\Cref{eq:nmatch}. In the following, we provide additional technical details required to derive the exact optimization steps and compute $\nmatch$ in practice. 

Following~\cite[Sec. 3]{eisenberger2020deep}, we first introduce the following latent feature representation of a shape $\cXi$ that is used within each DeepShells layer
\begin{equation}
    \mFk_{\cXi}:=\biggl(\mPsik~,~\Sk\bigl(\mVi\bigr)~,~\mNk\biggr)\in\bbR^{m\times (k+6)}
\end{equation} 
where $\mPsik$ are the first $k$ eigenfunctions of the intrinsic Laplace-Beltrami operator on $\cXi$, corresponding to the smallest eigenvalues. The operator $\Sk$ is the smoothing map initially proposed in~\cite[Eq. (8)]{eisenberger2020smooth}. The matrix $\mNk$ denotes the outer normals of the (smoothed) input geometry. 

Overall, the resulting feature tensor $\mFk_{\cXi}$ yields a $(k+6)$-dimensional embedding per vertex $\mVi\in\bbR^{m\times 3}$, depending on the number of eigenfunctions $k$. In order to align two shapes in this embedding space, an affine transformation is proposed 
\begin{multline}
    \mFhk_{\cXi}\bigl(\mCk,\mtauk\bigr):=\biggl(\mPsik{\mCk}^\dagger~,~\Sk\bigl(\mVi\bigr)+\\\mPsik\mtauk~,~\mNhk\biggr)\in\bbR^{m\times (k+6)},
\end{multline} 
that deforms the input shape $\cXi$ in the $(k+6)$-dimensional embedding space. This deformation is parameterized with a functional map $\mCk\in\bbR^{k\times k}$~\cite{ovsjanikov2012functional} and displacement coefficients $\mtauk\in\bbR^{k\times 3}$. The outer normals of the deformed pose are denoted as $\mNhk$.

As we discussed in~\Cref{sec:architecture}, the correspondence task in our framework is fully specified by the optimal transport energy~\Cref{eq:ot}. To make the resulting update steps differentiable, an additional entropy regularization term is added to the energy
\begin{multline}
    E_\mathrm{match,reg}\bigl(\mF,\mG;\mPisoft\bigr)=E_\mathrm{match}\bigl(\mF,\mG;\mPisoft\bigr)+\\\lambda_\mathrm{ent}\sum_{i',j'}\mPisoft_{i',j'}\log \mPisoft_{i',j'}.
\end{multline}
This is a common approach that was initially proposed by the seminal work of Cuturi et al.~\cite{cuturi2013sinkhorn}. One compelling implication is that $E_\mathrm{match,reg}$ can be minimized efficiently with respect to the transport plans $\mPisoft\in[0,1]^{m\times n}$ through Sinkhorn's algorithm. Moreover, each individual update step of this algorithm is differentiable which makes it viable for standard gradient-based optimization. The resulting map $\nmatch$ is specified by the following alternating scheme of optimization steps
\begin{subequations}\label{eq:schemedeepshells}
\begin{align}\label{eq:schemedeepshellsa}
    \bigl(\mCk,\mtauk\bigr)\mapsto&\underset{\substack{\mPisoftk\in\\\cT(\cXi,\cXj)}}{\arg\min}E_\mathrm{match,reg}\biggl(\mFhk_{\cXi},\mFk_{\cXj};\mPisoftk\biggr),\\\label{eq:schemedeepshellsb}
   \mPisoftk\mapsto&~~\underset{\mCk,\mtauk}{\arg\min}~~E_\mathrm{match,reg}\biggl(\mFhk_{\cXi},\mFk_{\cXj};\mPisoftk\biggr),
\end{align}
\end{subequations}
Through this scheme, the minimization of the energy $E_\mathrm{match,reg}$ is decoupled into two separate update steps, each of which can be solved efficiently in closed form. The first expression is minimized via Sinkhorn's algorithm to obtain an optimal transport plan $\mPisoftk$ from the transportation polytope $\cT(\cXi,\cXj)$. The second update results in a standard linear least squares problem, see~\cite[Sec. 3]{eisenberger2020deep} for additional details. 
For the initial step, we replace the $(k+6)$-dimensional feature embeddings with the learned features $\mFi$ and $\mFj$ produced by the DiffusionNet backbone. The map $\nmatch$ then alternates between the minimization steps~\Cref{eq:schemedeepshellsa} and~\Cref{eq:schemedeepshellsb} while increasing the number of eigenfunctions $k$ after each step. The final outputs are defined as
\begin{subequations}\label{eq:outputsnmatch}
\begin{align}\label{eq:outputsnmatcha}
    \mPiij:=&\underset{\mPi}{\arg\min}~~~E_\mathrm{match}\biggl(\mFh^{(k_\mathrm{max})}_{\cXi},\mF^{(k_\mathrm{max})}_{\cXj}~;~\mPi\biggr)
    \\\label{eq:outputsnmatchb}
    \mVij:=&\mVi+\mPsi^{(k_\mathrm{max})}\mtau^{(k_\mathrm{max})},
    \\\label{eq:outputsnmatchc}
    \ell_\mathrm{match}^{(i,j)}:=&\sum_kE_\mathrm{match}\biggl(\mFhk_{\cXi},\mFk_{\cXj}~;~\mPisoftk\biggr).
\end{align}
\end{subequations}
The matches $\mPiij\in\{0,1\}^{m\times n}$, with ${\mPiij}\mathbf{1}_n=\mathbf{1}_m$, produced by $\nmatch$ are thereby the outputs of the final optimization layer $k_\mathrm{max}$. In practice, they are obtained as the hard nearest-neighbor assignment between the final obtained shape embeddings $\mFh^{(k_\mathrm{max})}_{\cXi}$ and $\mF^{(k_\mathrm{max})}_{\cXj}$. 

\begin{figure*}
\begin{subfigure}[b]{0.24\textwidth}
    \centering
    \includegraphics[width=\linewidth]{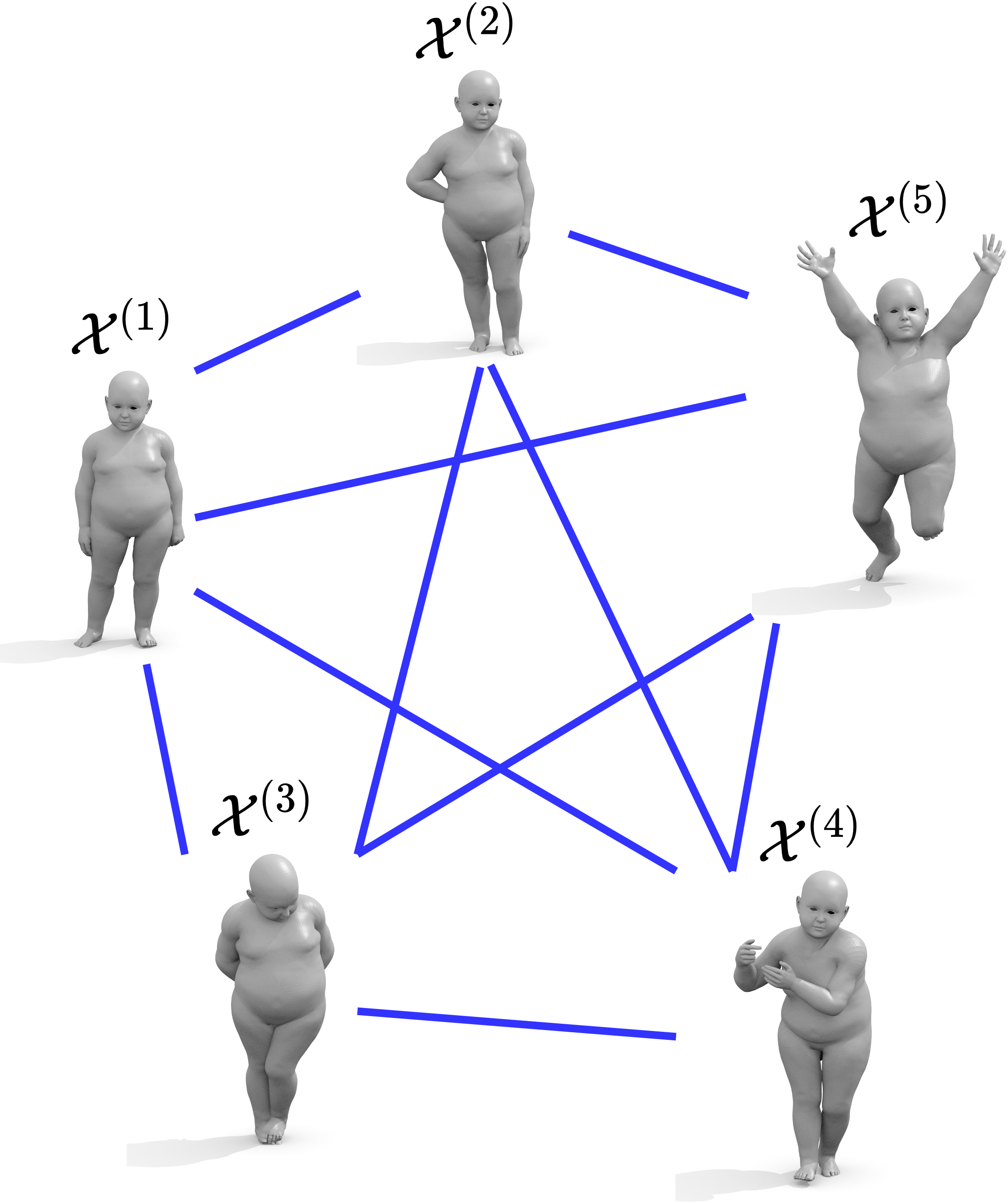}
    \label{fig:graph_type_a}
    \caption{Full graph.}
\end{subfigure}
\begin{subfigure}[b]{0.24\textwidth}
    \centering
    \includegraphics[width=\linewidth]{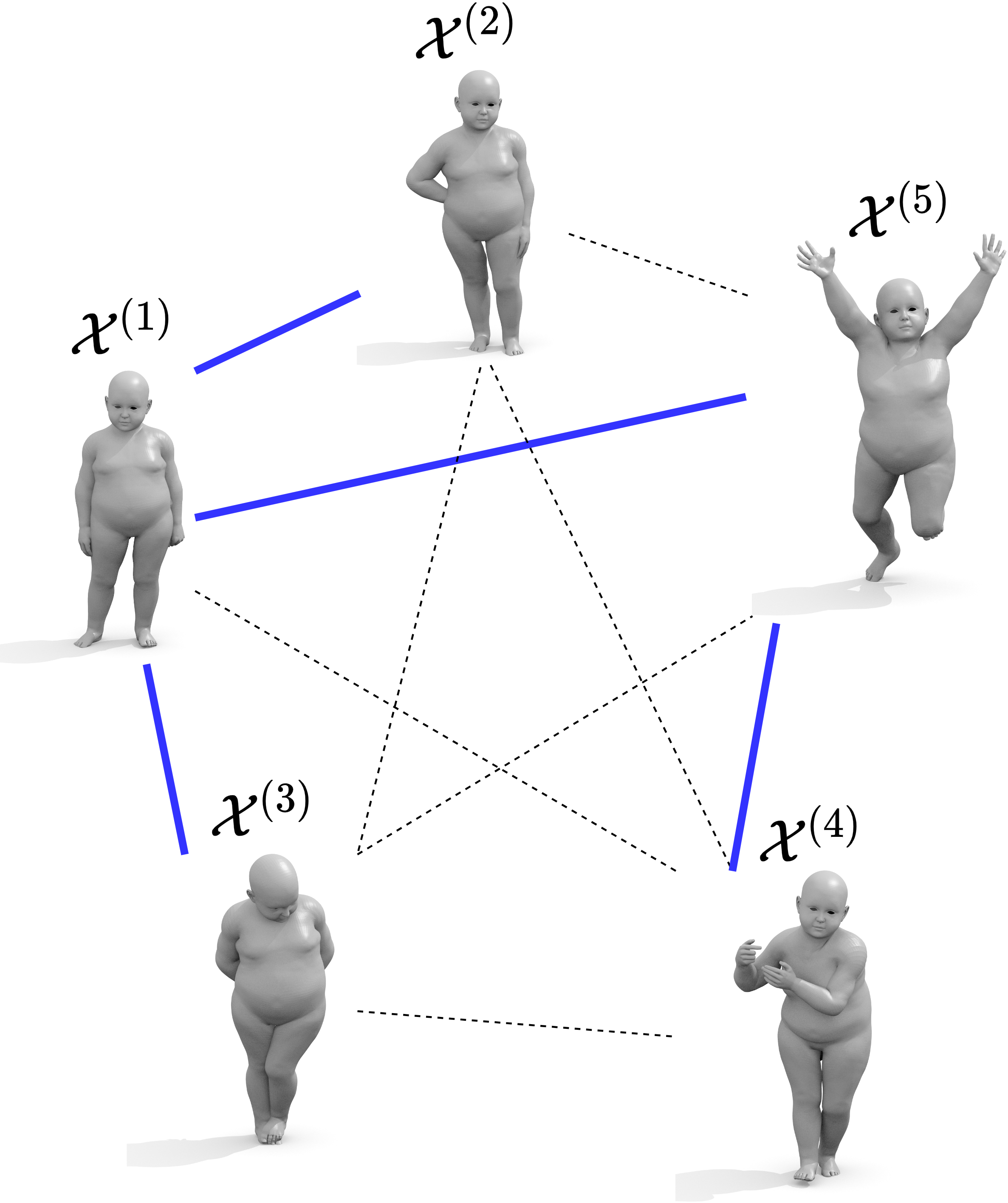}
    \label{fig:graph_type_b}
    \caption{MST graph.}
\end{subfigure}
\begin{subfigure}[b]{0.24\textwidth}
    \centering
    \includegraphics[width=\linewidth]{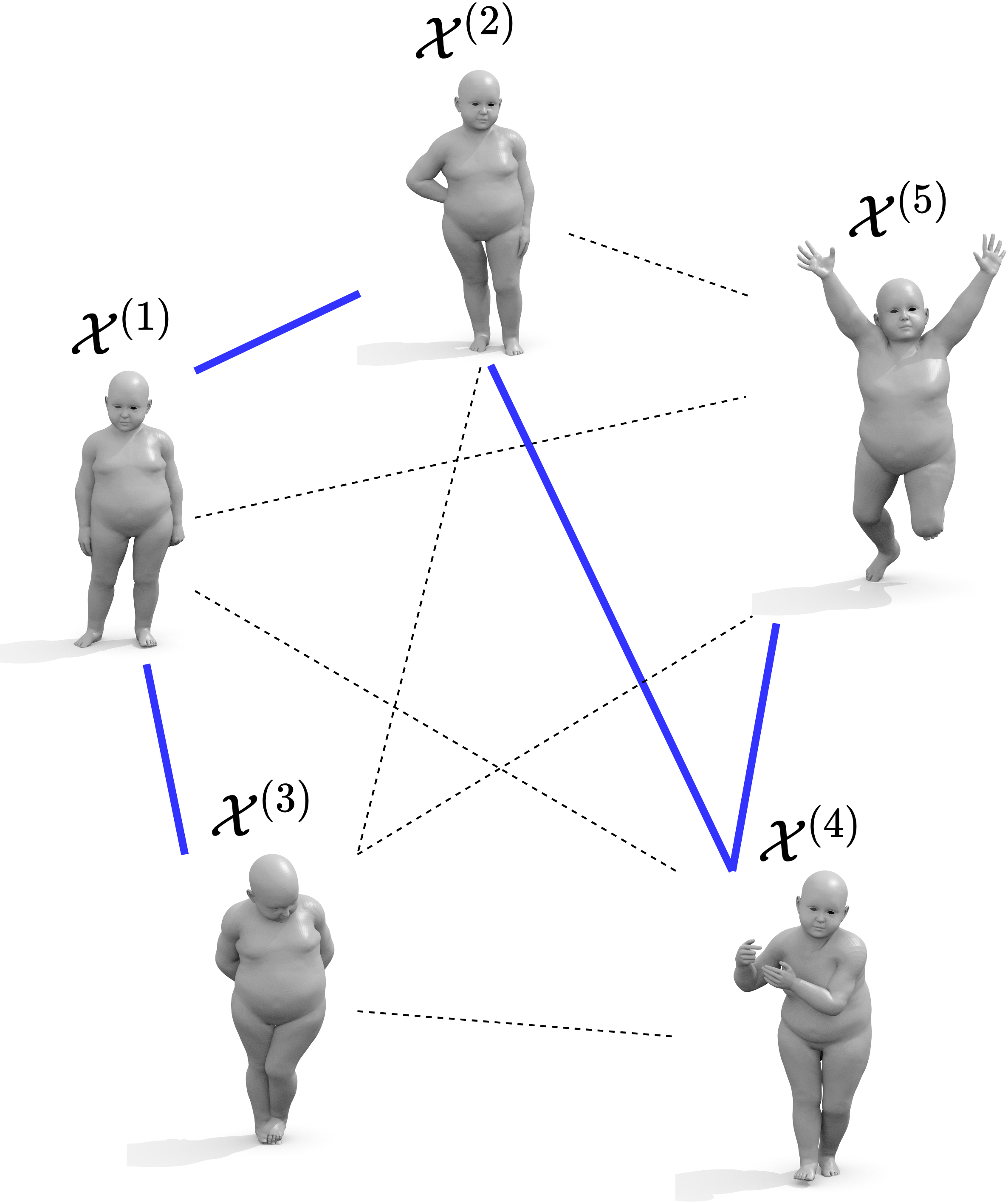}
    \label{fig:graph_type_c}
    \caption{TSP graph.}
\end{subfigure}
\begin{subfigure}[b]{0.24\textwidth}
    \centering
    \includegraphics[width=\linewidth]{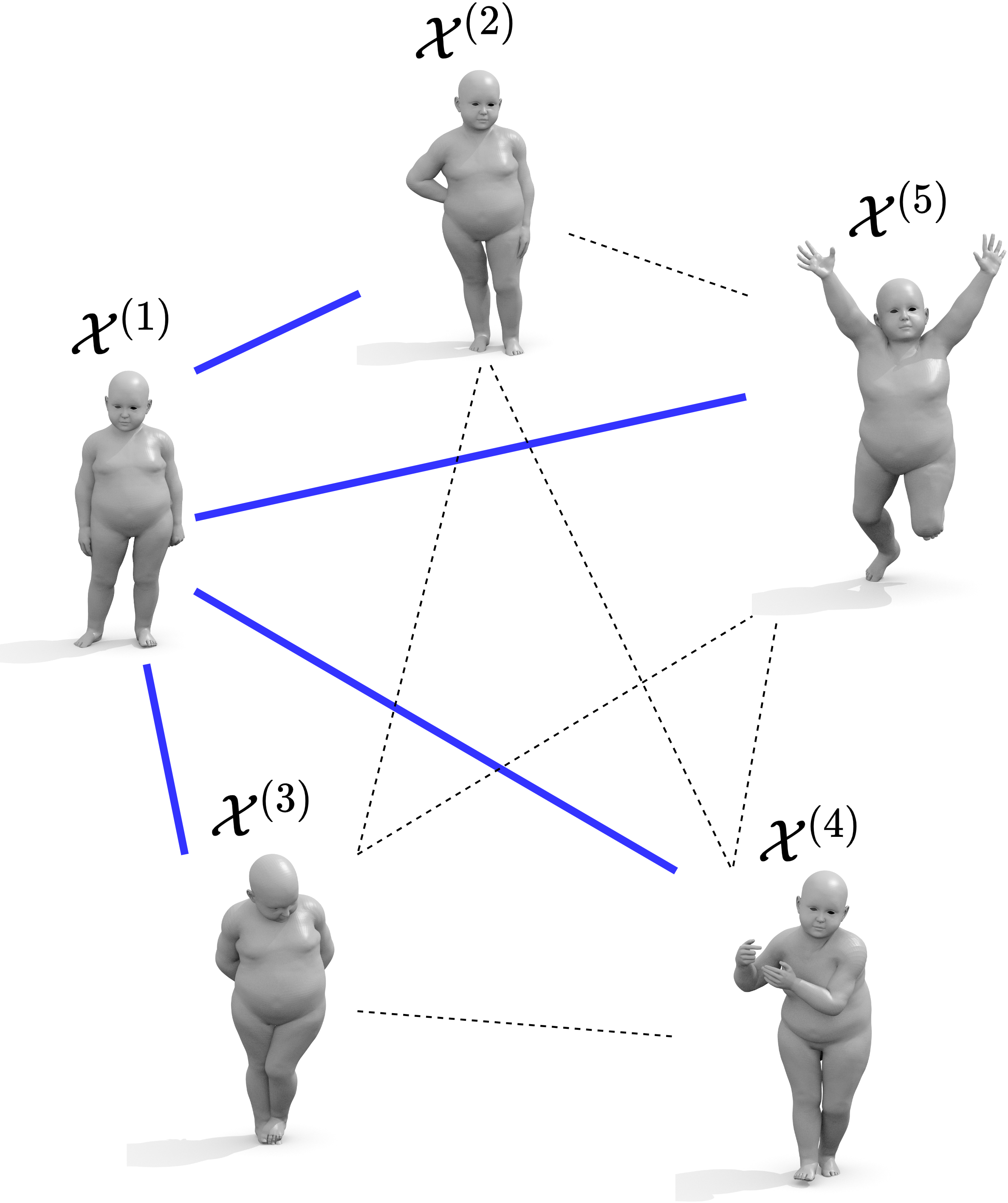}
    \label{fig:graph_type_d}
    \caption{Star graph.}
\end{subfigure}
\caption{An overview of different shape graph topologies $\cG$. We consider graphs that are (i) fully connected, (ii) minimal spanning trees, (iii) minimal Hamiltonian paths, specified by the traveling salesman problem and (iv) star graphs centered around one canonical pose. We provide a detailed discussion in~\Cref{app:graphtopologies}. Unless stated otherwise, the full graph (i) is the default for our approach.
}
\label{fig:graph_type}
\end{figure*}

\section{Shape graph topology}\label{app:graphtopologies}
We explore the following classes of graph topologies for the shape graph $\cG$, see also~\Cref{fig:graph_type} for a visualization:
\begin{enumerate}
    \item[i.] \textbf{Full graph}. The default setting for our method is the fully connected graph with the edge weights defined in~\Cref{eq:weights}.
    \item[ii.] \textbf{MST graph}. We consider the minimal spanning tree corresponding to the full graph $\cG$. This graph topology is a minimal choice, in the sense that it has the smallest total edge weight among all subgraphs of $\cG$ that span the set of nodes $\cS$.   
    \item[iii.] \textbf{TSP graph}. Based on the traveling salesman problem, we predict a Hamiltonian path of minimal total edge weight. This effectively defines an optimal ordering of the input set $\cS$.
    \item[iv.] \textbf{Star graph}. We define a complete, bipartite graph which connects one specific center node with all $N-1$ remaining nodes. The idea is to imitate template-based shape matching methods such as~\cite{bhatnagar2020loopreg,groueix20183dcoded} where all training poses are matched to a canonical shape. 
\end{enumerate}
Unless stated otherwise, we use the fully connected graph (i) by default in our experiments in~\Cref{sec:experiments}.
Choosing the number of retained edges is generally subject to a trade-off between accuracy and efficiency. Using (i) all $\frac{N(N-1)}{2}$ edges often yields the most accurate matching $\mPiijmult$, since this leads to the shortest possible path lengths in~\Cref{eq:pirefinea}. Nevertheless, the sparse graph topologies (ii)-(iv) might be preferable for specific applications. All three definitions (ii)-(iv) specify variants of spanning trees with exactly $N-1$ edges. This means, that the memory complexity for storing the graph, as well as the full query runtime cost is in $\mathcal{O}(N)$, see~\Cref{app:computationalcomplexity} for a cost analysis. Thus, they are more suitable for very large training sets or in scenarios where computational resources are scarce. See~\Cref{sec:comparisongraphtopologies} in the main paper for an empirical comparison of the different topologies (i)-(iv).

\paragraph{Discussion} Each of the variants (ii)-(iv) has interesting properties that might give rise to potential new avenues of applications in future work: (ii) Removing the (k-1) largest edges in the MST graph yields a subdivision of $\cG$ into k optimal clusters (MST clustering). (iii) The TSP graph orders the input shapes into a sequence, i.e., predicts a canonical ordering. (iv) Choosing an optimal star graph automatically selects one of the shapes in the collection $\cS$ as the canonical pose. By comparing the set of all possible star graphs, we can in principle rank all input poses in terms of how representative they are of the underlying shape manifold. 

\begin{table*}
\resizebox{0.90\linewidth}{!}{
\begin{tabular}{l|cccc}
\toprule[0.2em]
& $\#\mathrm{pairs}=10^2$ & $20^2$ & $50^2$ & $100^2$\\
\toprule[0.2em]
Epoch training time ($s$) & $37.85\pm2.48$ & $174.82\pm8.31$ & $921.23\pm25.56$ & $4192.91\pm142.03$ \\
Graph construction ($s$) & $36.04\pm1.95$ & $157.30\pm7.04$ & $893.23\pm\pz9.05$ & $3814.96\pm\pz44.48$ \\
\midrule[0.05em]
Required RAM ($\mathrm{GB}$) & $3.57\pm0.02$ & $3.96\pm0.04$ & $5.81\pm0.28$ & $9.02\pm1.73$ \\
\midrule[0.1em]
\end{tabular}
}
\vspace{5pt}
\centering
\caption{\textbf{Empirical training cost.} We quantify the computation cost of our pipeline for different training set sizes. For a given number of shapes $N=|\cS|$, one epoch consists of $\#\mathrm{pairs}=N^2\in\{10^2,\dots,100^2\}$ optimization steps that each match a pair of shapes $\cXi,\cXj\in\cS$.
}
\label{tab:computationcosttrain}

\end{table*}
\begin{table*}
\resizebox{0.90\linewidth}{!}{
\begin{tabular}{l|lcccc}
\toprule[0.2em]
& & $\#\mathrm{pairs}=10^2$ & $20^2$ & $50^2$ & $100^2$ \\
\toprule[0.2em]
Total query & Full graph & $41.00\pm3.32$ & $166.91\pm9.54$ & $1077.06\pm28.35$ & $4370.36\pm115.00$ \\
time ($s$) & MST graph & $\pz3.62\pm0.29$ & $\pz\pz8.10\pm0.36$ & $\pz\pz22.48\pm\pz0.90$ & $\pz\pz50.98\pm\pz\pz2.57$ \\
\midrule[0.05em]
Graph storage & Full graph & $10.66\pm0.00$ & $42.64\pm0.01$ & $266.47\pm0.08$ & $1065.89\pm0.33$ \\
($\mathrm{MB}$) & MST graph & $\pz0.96\pm0.00$ & $\pz2.03\pm0.00$ & $\pz\pz5.22\pm0.00$ &  $\pz\pz10.55\pm0.00$ \\
\bottomrule[0.1em]
\end{tabular}
}
\vspace{5pt}
\centering
\caption{\textbf{Total query cost.} We report the computation cost of our pipeline at test time. Note, that these results only apply to the specific setting where all pairs of a given set of shapes are queried and the graph $\cG$ is \emph{precomputed}. Under these circumstances, the MST graph proves to be superior as its computation cost increases linearly $\mathcal{O}(N)$ in the number of shapes $N$. For pairwise matching at test time, or when the graph $\cG$ needs to be extracted on the fly, the advantages of MST are less prominent.
}
\label{tab:computationcosttest}

\end{table*}

\section{Empirical computation cost}\label{app:computationalcomplexity}

\paragraph{Training} We empirically measure the computation cost of our full pipeline. To this end, we choose a training set of $\{10^2,20^2,50^2,100^2\}$ pairs from the SURREAL~\cite{varol2017learning} dataset with a fixed mesh resolution of $6890$ vertices, which is common for SMPL~\cite{SMPL:2015} meshes. The resulting training runtime and memory costs are summarized in~\Cref{tab:computationcosttrain}. Averaged over all samples, our model takes around $\approx0.4s$ per training pair. The majority of the cost for constructing the graph $\cG$ results from querying all sample pairs, which is equivalent to the epoch training cost minus the backward pass. The remaining cost stems from precomputing the concatenated, pairwise matches as discussed in~\Cref{app:trainingdetails}.

\paragraph{Query time} We additionally compare the required cost for querying our model. Aside from our main pipeline, we also consider the sparse `MST' graph type introduced in~\Cref{app:graphtopologies}. The resulting costs are summarized in~\Cref{tab:computationcosttest}. For a given set of query shapes, one has to distinguish whether the graph $\cG$ is precomputed or needs to be predicted on the fly. In the latter case, an additional cost of constructing the graph is added, see the second row of~\Cref{tab:computationcosttrain}. Notably, this cost does not depend on the test graph topology. This also means that the main computational advantages of the MST graph are less prevalent when no offline graph precomputation is possible, e.g., for a pair of unseen test poses. When the unseen pose is supposed to be added to an existing graph in an online fashing, only $N$ pairs between the old training set and the new pose need to be computed. This, however, again entails the same cost for either the full graph or MST. On the other hand, MST is much faster for precomputed graphs. Also, the storage cost of the MST graph is always more efficient than the dense `full' setting. This makes sparse graph topologies relevant when memory is limited or for very large training sets, since the required memory of the full graph grows quadratically $\mathcal{O}(N^2)$ with the number of training shapes $N=|\cS|$.

\section{Qualitative results}\label{app:qualitative}

\begin{figure*}
    \centering
    \begin{overpic}
    [width=\linewidth]{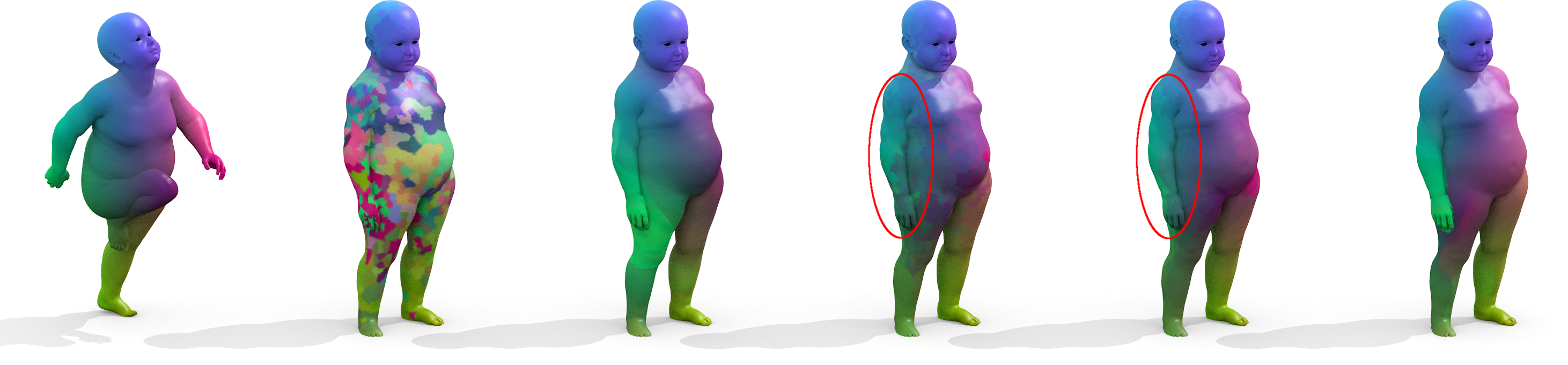}
    \put(5.5,25){Target}
    \put(21,25){UDM~\cite{cao2022unsupervised}}
    \put(39,25){DS~\cite{eisenberger2020deep}}
    \put(55,25){NM~\cite{eisenberger2021neuromorph}}
    \put(71,25){SyNoRiM~\cite{huang2022multiway}}
    \put(91,25){Ours}
    \end{overpic}
    \includegraphics[width=\linewidth]{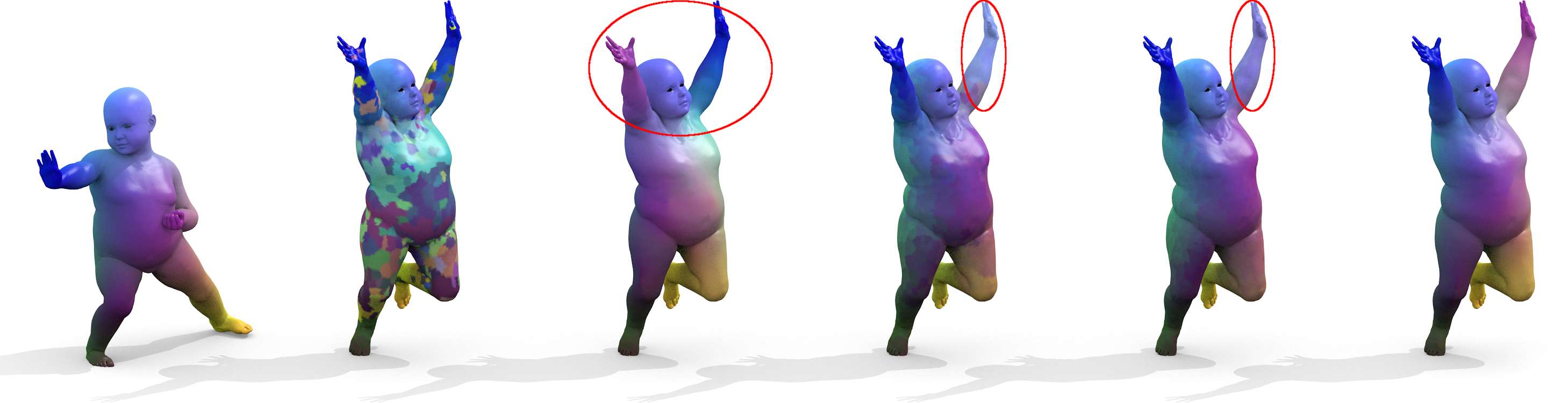}
    \includegraphics[width=\linewidth]{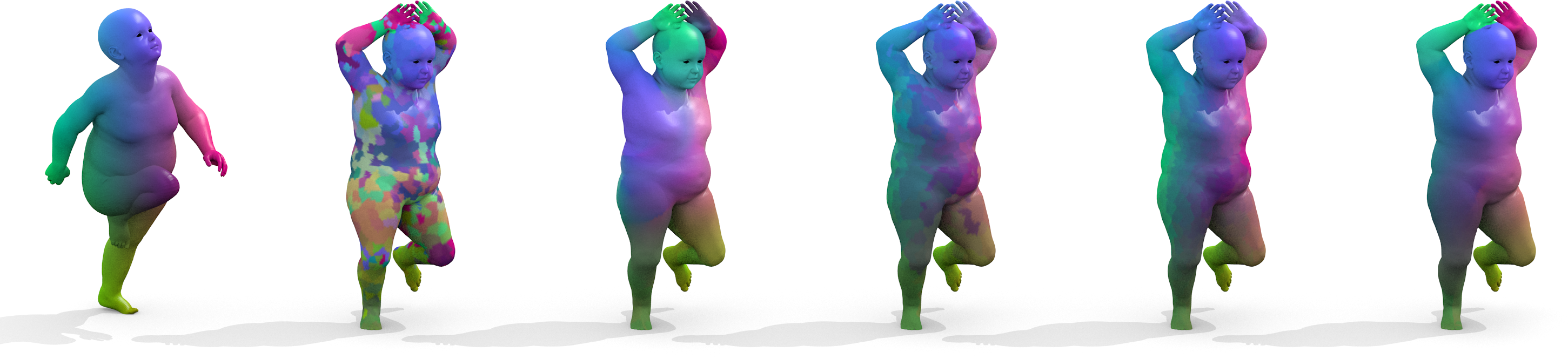}
    \caption{\textbf{Topological noise qualitative.} We compare the quality of the predicted correspondences on three pairs from TOPKIDS~\cite{laehner2016shrec}, corresponding to the quantitative results from~\Cref{fig:geodesic_shrec_top}. All three pairs are corrupted by topological noise in places of self-contact, e.g., where the child's arms touch its upper body or head.
    }
    \label{fig:app_qual_topo}
\end{figure*}

\begin{figure*}
    \centering
    
    \vspace{9pt}
    \begin{overpic}
    [width=\linewidth]{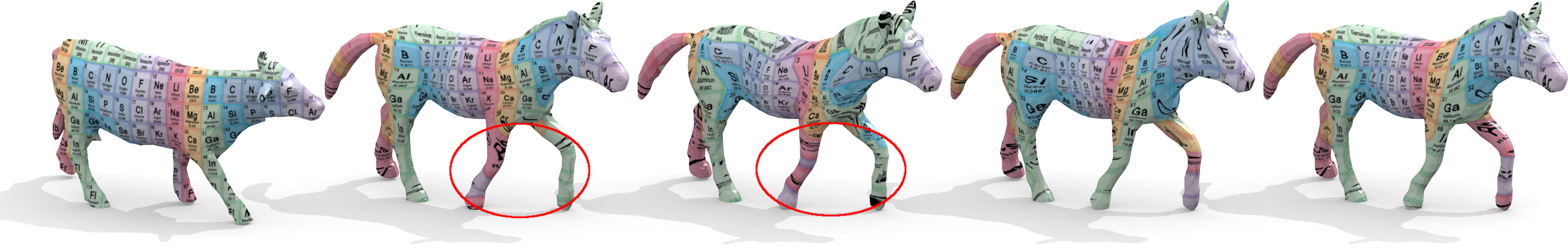}
    \put(7,19){Target}
    \put(26.5,19){DS~\cite{eisenberger2020deep}}
    \put(47,19){NM~\cite{eisenberger2021neuromorph}}
    \put(65,19){SyNoRiM~\cite{huang2022multiway}}
    \put(87,19){Ours}
    \end{overpic}
    
    \vspace{9pt}
    \includegraphics[width=\linewidth]{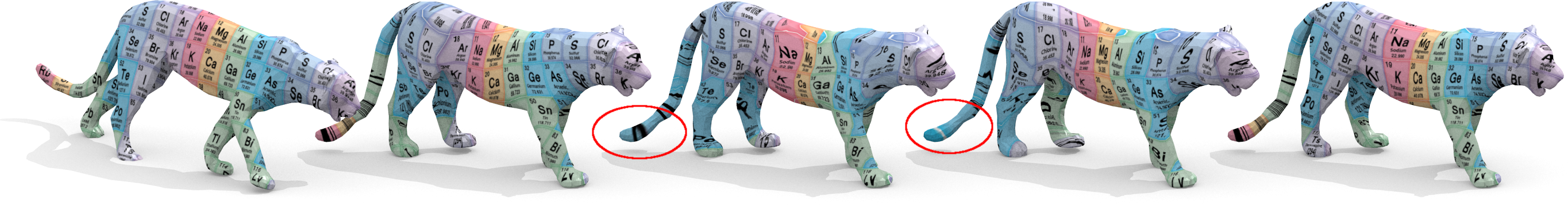}
    
    \vspace{9pt}
    \includegraphics[width=\linewidth]{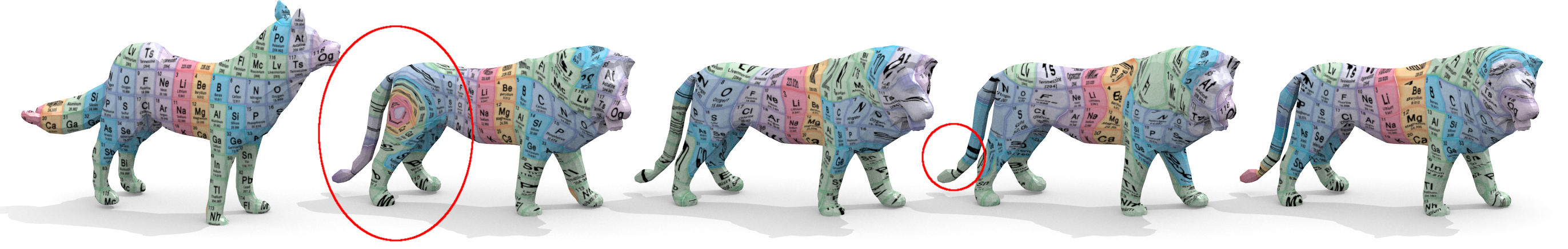}
    
    \vspace{9pt}
    \includegraphics[width=\linewidth]{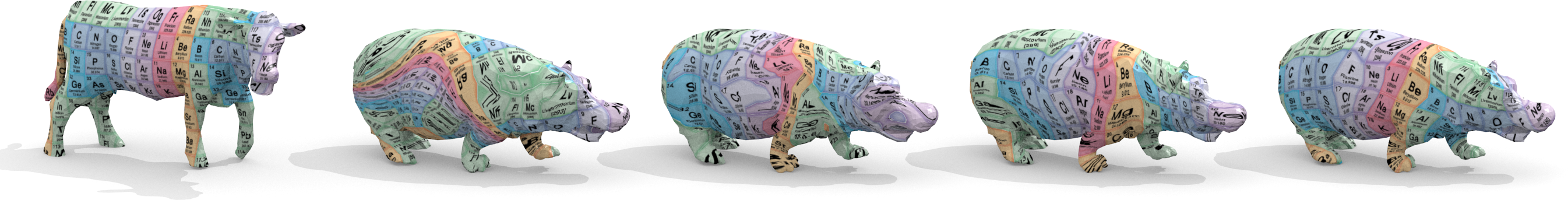}
    
    \vspace{9pt}
    \includegraphics[width=\linewidth]{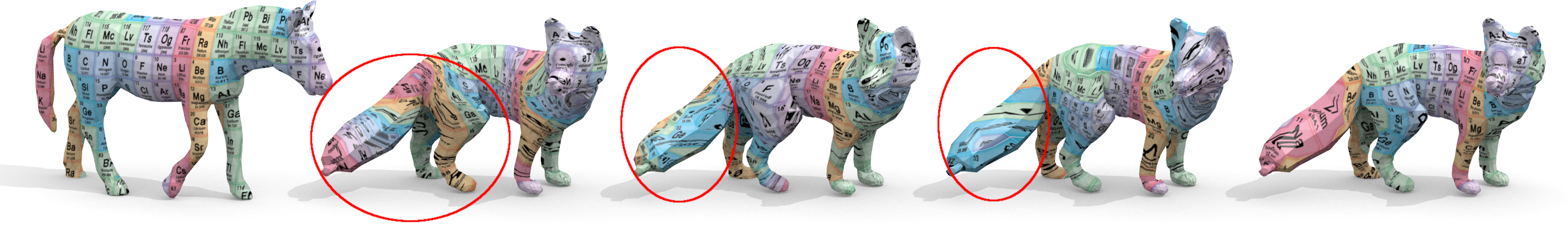}
    \caption{\textbf{Inter-class texture transfers.} We assess the map smoothness of several baseline approaches, in comparison to our proposed method. The five sample pairs are taken from the SMAL~\cite{zuffi20173d} test set corresponding to our benchmark comparison in~\Cref{fig:interclass}. In each case, the obtained matches are visualized via a texture map.
    }
    \label{fig:app_qual_interclass}
\end{figure*}

For a more complete picture, we provide several additional qualitative comparisons.
~\Cref{fig:app_qual_topo} and~\Cref{fig:app_qual_interclass} show results corresponding to the benchmark comparisons from~\Cref{sec:topologicalmatching} and~\Cref{sec:interclassmatching} of the main paper. 

\end{document}

%% file: commands.tex
\newcommand{\mX}{\mathbf{X}}
\newcommand{\mY}{\mathbf{Y}}
\newcommand{\mD}{\mathbf{D}}
\newcommand{\mV}{\mathbf{V}}
\newcommand{\mVi}{\mV^{(i)}}
\newcommand{\mVj}{\mV^{(j)}}
\newcommand{\mVij}{\mV^{(i,j)}}
\newcommand{\mVji}{\mV^{(j,i)}}
\newcommand{\mN}{\mathbf{N}}
\newcommand{\mNi}{\mN^{(i)}}
\newcommand{\mNj}{\mN^{(j)}}
\newcommand{\mNk}{\mN^{(k)}}
\newcommand{\mF}{\mathbf{F}}
\newcommand{\mFi}{\mF^{(i)}}
\newcommand{\mFj}{\mF^{(j)}}
\newcommand{\mFk}{\mF^{(k)}}
\newcommand{\mFh}{\hat{\mF}}
\newcommand{\mFhk}{\mFh^{(k)}}
\newcommand{\mG}{\mathbf{G}}
\newcommand{\mT}{\mathbf{T}}
\newcommand{\mTi}{\mT^{(i)}}
\newcommand{\mTj}{\mT^{(j)}}
\newcommand{\cX}{\mathcal{X}}
\newcommand{\cXi}{\cX^{(i)}}
\newcommand{\cXj}{\cX^{(j)}}
\newcommand{\cS}{\mathcal{S}}
\newcommand{\cG}{\mathcal{G}}
\newcommand{\mPi}{\mathbf{\Pi}}
\newcommand{\mPiij}{\mPi^{(i,j)}}
\newcommand{\mPiijmult}{\mPiij_\mathrm{mult}}
\newcommand{\mPiji}{\mPi^{(j,i)}}
\newcommand{\mPisoft}{\mathbf{\tilde{\Pi}}}
\newcommand{\mPik}{\mPi^{(k)}}
\newcommand{\mPisoftk}{\mPisoft^{(k)}}
\newcommand{\mPsi}{\mathbf{\Psi}}
\newcommand{\mPsik}{\mPsi^{(k)}}
\newcommand{\Sk}{S^{(k)}}
\newcommand{\mC}{\mathbf{C}}
\newcommand{\mCk}{\mC^{(k)}}
\newcommand{\mtau}{\mathbf{\tau}}
\newcommand{\mtauk}{\mtau^{(k)}}
\newcommand{\mNh}{\hat{\mN}}
\newcommand{\mNhk}{\mNh^{(k)}}
\newcommand{\mL}{\mathbf{L}}
\newcommand{\mS}{\mathbf{S}}
\newcommand{\mM}{\mathbf{M}}
\newcommand{\mLambda}{\mathbf{\Lambda}}

\newcommand{\vf}{\mathbf{f}}

\newcommand{\cY}{\mathcal{Y}}
\newcommand{\cT}{\mathcal{T}}
\newcommand{\bbR}{\mathbb{R}}

\newcommand{\nfeat}{\Phi_\mathrm{feat}}
\newcommand{\nmatch}{\Phi_\mathrm{match}}

\newcommand{\pz}{\phantom{0}}
\newcommand{\yc}{\cellcolor[HTML]{FFFFC0}}